\documentclass[runningheads]{llncs}

\usepackage{graphicx}
\usepackage{cite}
\usepackage{amsmath,amssymb,amsfonts}
\usepackage{algorithmic}
\usepackage{textcomp}
\usepackage{xcolor}
\usepackage{hyperref}
\usepackage{booktabs}
\usepackage{multirow}
\usepackage{enumitem}
\usepackage{tikz}
\usetikzlibrary{shapes,arrows,positioning}

\usepackage{listings}
\usepackage{array}
\newcolumntype{L}[1]{>{\raggedright\arraybackslash}p{#1}}
\newcolumntype{C}[1]{>{\centering\arraybackslash}p{#1}}

\usepackage{comment}

\begin{document}

%
%
\title{Architecting Agentic Communities using Design Patterns}

\subtitle{A Framework Grounded in ODP Enterprise Language Formalism}

%
%
\author{Zoran Milosevic\inst{1,2}\orcidID{0000-0002-1364-7423} \and
Fethi Rabhi\inst{1} \orcidID{0000-0001-8934-6259}}

\authorrunning{Z. Milosevic et al.}

\institute{School of Computer Science and Engineering, University of New South Wales, Sydney, Australia\\
\email{(z.milosevic,f.rabhi)@unsw.edu.au}
\and
Deontik, Australia\\
\email{zoran@deontik.com}}

\maketitle

%
%
\begin{abstract}

The rapid evolution of Large Language Models (LLM) and subsequent Agentic AI 
technologies requires systematic architectural guidance 
for building sophisticated, production-grade systems. This paper presents an 
approach for architecting such systems using design patterns derived from 
enterprise distributed systems standards, formal methods, and industry practice. 
We classify these patterns into three tiers: LLM Agents (task-specific automation), 
Agentic AI (adaptive goal-seekers), and Agentic Communities (organizational 
frameworks where AI agents and human participants coordinate through formal 
roles, protocols, and governance structures). We focus on Agentic Communities---
coordination frameworks encompassing LLM Agents, Agentic AI entities, and 
humans---most relevant for enterprise and industrial applications. 
Drawing on established coordination principles from distributed systems, we 
ground these patterns in a formal framework that specifies collaboration 
agreements where AI agents and humans fill roles within governed 
ecosystems. This approach provides both practical guidance and formal 
verification capabilities, enabling expression of organizational, legal, and 
ethical rules through accountability mechanisms that ensure operational and verifiable governance 
of inter-agent communication, negotiation, and intent modeling. We validate 
this framework through a clinical trial matching case study. Our goal is to 
provide actionable guidance to practitioners while maintaining the formal 
rigor essential for enterprise deployment in dynamic, multi-agent ecosystems.

\end{abstract}

%
%
\section{Introduction}
\label{sec:introduction}

The emergence of Large Language Models (LLMs) has significantly transformed 
software development, enabling creation of \textit{LLM Agents}---task-specific entities leveraging 
LLMs for narrow automation tasks. Subsequently, advances in reasoning capabilities, 
tool integration, and prompting techniques led to \textit{Agentic AI}---software entities 
with genuine agency combining autonomy with goal-directed reasoning, able to perceive environments, formulate plans, and adapt strategies 
dynamically \cite{wang2024survey}. Building upon these capabilities gives rise to, what we refer to as, \textit{Agentic Communities}---
collaboration frameworks where multiple LLM Agents, Agentic AI 
entities, and human actors collaborate through structured protocols to achieve 
common objectives beyond individual agent capabilities. Unlike traditional software agents with pre-deterministic behaviors, 
or LLM Agents executing assigned tasks without independent goal formulation, Agentic 
AI represents a fundamental departure through autonomous goal pursuit, while Agentic 
Communities enable coordinated multi-participant systems with explicit and operational governance architecture essential for enterprise and industrial deployment.

However, this shift from deterministic to agentic behavior introduces significant architectural challenges. Practitioners face questions about when to employ single autonomous agents 
versus coordinated multi-agent systems, how to ensure accountability in 
multi-agent autonomous decision-making, how to integrate human oversight and expertise, 
and how to compose patterns effectively while maintaining governance and 
compliance. These challenges require systematic architectural frameworks 
grounded in well-proven and rigorous methods. The practical need for such frameworks is evidenced by emerging industrial implementations in asset-intensive industries. For example, production deployments of Multi-Agent Generative Systems incorporating formal governance and 
accountability mechanisms have demonstrated measurable operational improvements 
in mining, oil \& gas, and manufacturing sectors~\cite{xmpro_mags2025}, 
validating the practical viability of formal accountability in safety-critical 
autonomous systems.

While design patterns have proven invaluable in traditional software engineering \cite{gamma1995design} and distributed systems \cite{hohpe2003enterprise,rabhi2003DPs}, their application to agentic AI systems is limited to simple ones such as ReAct~\cite{yao2023react}, Tool Use~\cite{schick2023toolformer}, 
Planning~\cite{huang2022language}, and Reflection~\cite{shinn2023reflexion}. Current pattern catalogues either focus narrowly on specific LLM prompting techniques or lack the formal rigor necessary for enterprise-grade systems with stringent accountability requirements. Moreover, they rarely address the reality that enterprise systems must 
coordinate both AI agents and human participants within governed frameworks.

To address this gap, this paper provides a systematic approach to classifying different types of LLM-powered 
entities---LLM Agents, Agentic AI, and Agentic Communities---and identifying patterns that are relevant for each of these types. The paper uses ISO Open Distributed Processing (ODP) Enterprise Language (EL) community formalism \cite{IS15414} as a foundation for describing coordination challenges in Agentic Communities, including the relationship between intent and obligations, critical for reasoning about accountability when 
autonomous agents and humans collaborate within governed frameworks.

The remainder of this paper is organized as follows. Section \ref{sec:background} provides background about LLM powered agents and Agentic AI, and introduces  ODP EL community formalism. Section \ref{sec:catalogue} presents our comprehensive pattern catalogue with 
categories, classification methodology, and key insights. Section \ref{sec:design-methodology} presents a design pattern methodology adapted to agentic AI systems. Section \ref{sec:demonstration} validates the framework through a clinical trial matching case study. Section \ref{sec:formal-analysis} discusses the formal aspects of our architecture approach enabling rigorous operational and governance verification and implementations. Section \ref{sec:related-work} discusses related work, and Section \ref{sec:conclusion} provides concluding remarks and outlines future research directions.

\section{Background}
\label{sec:background}

\subsection{LLM Agent, Agentic AI and Agentic Community}
\label{subsec:three-tier}

We introduce a three-tier classification framework that builds upon and refines the conceptual taxonomy proposed by Sapkota et al. \cite{Sapkota_2026}. While their work distinguishes AI Agents (task-specific automation) from Agentic AI (systems with multi-agent collaboration and coordinated autonomy), we decompose their ``Agentic AI'' category into two distinct levels to address different architectural requirements: individual agents with genuine agency, and multi-participant coordination frameworks, including both Agentic AI and human participants. This refinement enables precise pattern selection based on whether systems require single autonomous agents or coordinated communities with formal governance.

\textbf{Key Terminology---Autonomy vs. Agency}: Before introducing our classification, we clarify two foundational concepts. \textit{Autonomy} denotes the ability to operate independently without constant external supervision---making decisions and taking actions without requiring continuous human control. \textit{Agency}, however, represents a richer concept encompassing autonomy plus intentionality, goal-directedness, and adaptive behavior \cite{wooldridge2009introduction}. An autonomous system operates independently; an agentic entity independently pursues goals through contextual reasoning, strategic planning, and adaptive action-taking. This distinction underpins our classification: LLM Agents exhibit task-scoped autonomy (independent operation within defined parameters) while Agentic AI exhibits genuine agency (autonomous goal pursuit with adaptive reasoning).

\textbf{LLM Agent}\footnote{Sapkota et al. \cite{Sapkota_2026} use the term ``AI Agents'' for this category; we use ``LLM Agent'' throughout to emphasize the LLM foundation that distinguishes these from traditional rule-based agents.}: LLM-powered software entities executing specific tasks autonomously within controlled environments. While capable of independent operation, LLM Agents lack agency---they execute assigned tasks without independent goal formulation or adaptive strategy adjustment. LLM Agents function as building blocks and utility components within larger systems. Examples include data validators, extractors, format converters, and filtering agents. An LLM Agent capability can be considered a simplified abstraction of an Agentic AI capability, focusing on task execution without the cognitive sophistication required for genuine agency.

\textbf{Agentic AI}: LLM-powered and other entities possessing genuine agency---the capability for autonomous goal pursuit, contextual reasoning, strategic planning, and adaptive behavior. Unlike LLM Agents, Agentic AI independently determines both what goals to pursue and how to achieve them, adapting strategies based on environmental context and past experience. Agentic AI represents the cognitive foundation enabling genuine autonomous behavior. Examples include ReAct \cite{yao2023react} agents with reasoning 
traces, hierarchical planning agents \cite{huang2022language}, reflexive learners that 
improve through self-critique \cite{shinn2023reflexion}, and metacognitive systems 
\cite{wang2024survey}.

\textbf{Agentic Community}: Coordination frameworks where multiple participants---including LLM Agents, Agentic AI software entities, and human actors---collaborate through structured protocols and shared infrastructure. Agentic Communities exhibit emergent behaviors beyond individual participants through multi-agent collaboration, dynamic task decomposition, and persistent shared memory. In such a framework, human actors function as coordinated participants, supervisors, or peer collaborators rather than external observers. Examples include multi-agent workflows, orchestration platforms, negotiation frameworks, and human-AI collaborative systems. This category encompasses and extends aspects of Sapkota et al.'s ``Agentic AI'' by adding explicit human integration and, importantly, formal accountability governance specification leveraging ODP-EL community formalism (Section \ref{subsec:odp-foundation-intro}).

Note that Agentic Communities may not always involve human users directly, but there must be traceability from community actions to the parties ultimately responsible for their effects---ensuring accountability in autonomous systems. Further, communities can model organizational aspects of a single organization (roles, capabilities, rules) or capture cross-organizational contracts and workflows, for which federations (a specialized community type) can be used. Section~\ref{subsec:odp-foundation-intro} details the ODP-EL formal foundation 
for Agentic Communities.

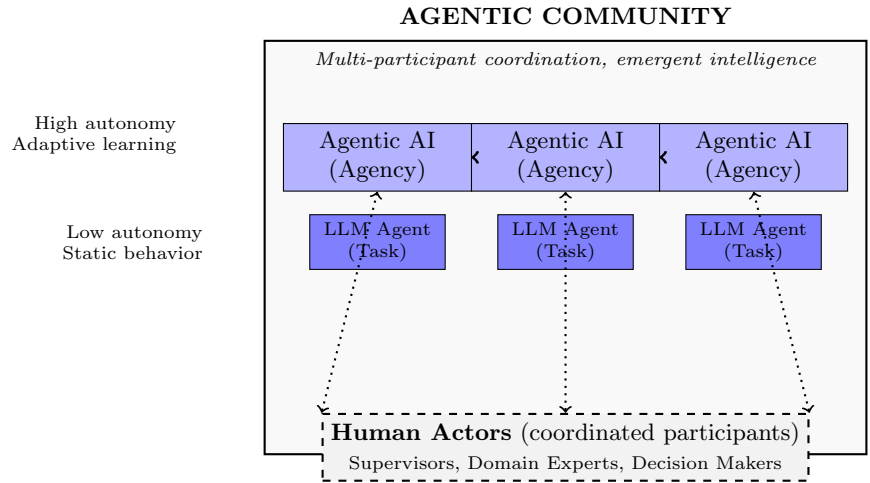
\begin{figure}[t]
\centering
\begin{tikzpicture}[
  box/.style={rectangle, draw, thick, minimum width=3.5cm, minimum height=1cm, align=center},
  smallbox/.style={rectangle, draw, minimum width=2.5cm, minimum height=0.8cm, align=center, font=\small},
  tinybox/.style={rectangle, draw, minimum width=1.8cm, minimum height=0.6cm, align=center, font=\scriptsize},
  human/.style={rectangle, draw, thick, dashed, minimum width=6cm, minimum height=0.8cm, align=center, fill=gray!10}
]

\node[box, minimum width=8cm, minimum height=5.5cm, fill=gray!5] (community) at (0,0) {};
\node[above=0.1cm of community.north, font=\bfseries] {AGENTIC COMMUNITY};
\node[below=0.05cm of community.north, font=\scriptsize\itshape] {Multi-participant coordination, emergent intelligence};

\node[smallbox, fill=blue!30] (agent1) at (-2.5,1.2) {Agentic AI\\(Agency)};
\node[smallbox, fill=blue!30] (agent2) at (0,1.2) {Agentic AI\\(Agency)};
\node[smallbox, fill=blue!30] (agent3) at (2.5,1.2) {Agentic AI\\(Agency)};

\node[tinybox, fill=blue!50, below=0.3cm of agent1] (task1) {LLM Agent\\(Task)};
\node[tinybox, fill=blue!50, below=0.3cm of agent2] (task2) {LLM Agent\\(Task)};
\node[tinybox, fill=blue!50, below=0.3cm of agent3] (task3) {LLM Agent\\(Task)};

\node[human, below=0.4cm of agent2.south, yshift=-0.6cm] (humans) at (0,-1.2) {
  \textbf{Human Actors} (coordinated participants)\\
  \scriptsize Supervisors, Domain Experts, Decision Makers
};

\draw[<->, dotted, thick] (agent1) -- (agent2);
\draw[<->, dotted, thick] (agent2) -- (agent3);
\draw[<->, dotted, thick] (agent1.south) -- (humans.north west);
\draw[<->, dotted, thick] (agent2.south) -- (humans.north);
\draw[<->, dotted, thick] (agent3.south) -- (humans.north east);

\node[left=1.3cm of task1, align=right, font=\scriptsize] {Low autonomy\\Static behavior};
\node[left=1.3cm of agent1, align=right, font=\scriptsize, yshift=0.3cm] {High autonomy\\Adaptive learning};

\end{tikzpicture}
\caption{Three types of LLM-powered entities and their relationships. Agentic Communities 
coordinate heterogeneous participants including LLM Agents, Agentic AI systems, 
and human actors through structured protocols, shared infrastructure and agreed and computable governance specification.}
\label{fig:hierarchy}
\end{figure}

These three types are illustrated in Figure \ref{fig:hierarchy}, noting increasing levels of autonomy, and emergent behavior, with Agentic Communities coordinating heterogeneous participants including both artificial and human agents.

\begin{table}[t]
\centering
\caption{Three-Tier Classification: Key Characteristics}
\label{tab:three-tier-characteristics}
\small
\begin{tabular}{@{}llll@{}}
\toprule
\textbf{Characteristic} & \textbf{LLM Agent} & \textbf{Agentic AI} & \textbf{Agentic Community} \\
\midrule
Autonomy Level & Limited & High & Distributed \\
Decision-Making & Prompt-driven & Context-aware & Collective \\
Scope & Specific tasks & Complex tasks & Multi-participant \\
Learning & Static & Continuous & Community-level \\
Goal Pursuit & Executes tasks & Formulates goals & Collaborative \\
Participants & Single component & Single agent & Multiple (AI + human) \\
Formal Model & N/A & N/A & ODP-EL Community \\
\bottomrule
\end{tabular}
\end{table}

\subsection{Formal Foundation: ODP-EL Communities}
\label{subsec:odp-foundation-intro}

As introduced in Section \ref{subsec:three-tier}, Agentic Communities 
coordinate heterogeneous participants---LLM Agents, Agentic AI entities, and 
human actors---through structured protocols within governed frameworks. To 
enable rigorous specification of such multi-participant coordination with 
verifiable governance properties, we ground Agentic Communities structure, behaviour and governance rules, in the ISO 
Open Distributed Processing Enterprise Language (ODP-EL) standard (ISO/IEC 
15414) \cite{IS15414,linington2011odp}---an internationally recognized framework 
for specifying enterprise distributed systems with rigorous, formal based semantics.

This formal grounding enables our pattern-based architectural approach by 
providing mechanisms for expressing design patterns as ODP-EL community 
templates---specifications that combine practical architectural guidance with 
formal semantics. Through this integration, patterns instantiate as community 
specifications comprising roles (behavioral placeholders fillable by LLM Agents, 
Agentic AI, or human actors), normative constraints (obligations, permissions, 
prohibitions), and contracts (normative relationships binding roles together). 
This dual nature---patterns remaining accessible to practitioners while supporting 
formal verification---addresses the critical gap between proven architectural 
solutions and the verifiable accountability, traceable authority, and provable 
compliance required for enterprise deployment in regulated industries.

The ODP-EL framework, proven in complex enterprise distributed systems theory and practice, provides 
essential constructs that transform our design patterns into precise architecture specifications, including verifiable 
properties:

\textbf{(1) Patterns as Community Templates}: Many design patterns instantiate as 
ODP-EL community specifications with roles (behavioral placeholders fillable 
by LLM Agents, Agentic AI, or human actors), normative constraints (obligations, 
permissions, prohibitions), and contracts (normative relationships binding 
roles together). This mapping enables patterns to remain practical while 
supporting formal verification.

\textbf{(2) Deontic Governance}: Pattern compositions express governance through 
burden, permit, and embargo tokens (referred to as deontic tokens), which encapsulate 
obligations, permissions and prohibitions; this facilitates creating formal 
accountability chains traceable across all participants which hold these deontic 
tokens. Deontic tokens enable formal delegation of obligations between 
agents while maintaining complete traceability of delegation chains, ensuring that 
authority and responsibility can be systematically transferred without losing 
accountability. This token-based approach enables automated audit trail generation 
and runtime policy enforcement.

\textbf{(3) Verifiable Properties}: ODP-EL's machine-checkable specifications 
enable systematic verification that pattern implementations satisfy governance 
requirements---producing verified authority boundaries, complete accountability 
chains, and compliance guarantees essential for regulatory acceptance.

\textbf{(4) Human-AI Integration}: The formalism provides uniform treatment of 
computational and human agents within patterns, while maintaining distinct 
legal responsibilities---essential for enterprise deployment where human oversight 
remains mandatory.

\textbf{(5) Runtime Enforcement}: Pattern implementations generate audit trails 
automatically, detect policy violations before compliance incidents, and enable 
deployment with verifiable properties rather than testing-based assurance alone.

This integration of design patterns with ODP-EL formal semantics distinguishes 
our work from narrative pattern catalogues. Section \ref{sec:formal-analysis} 
demonstrates how patterns map to ODP-EL constructs, combining practical 
architectural guidance with formal rigor. Section \ref{subsec:formal-properties} 
proves verifiable safety and authority properties in the clinical trial matching 
system, validating that pattern-based designs can achieve production-grade 
deployment with regulatory compliance.

By grounding practical patterns in proven ISO standards, we provide architects 
with both implementable guidance and verifiable specifications---the combination 
essential for enterprise agentic AI systems.

\subsection{Contributions}
\label{subsec:contributions}

This paper makes the following contributions: (1) A classification framework distinguishing LLM Agents, Agentic AI, and Agentic Communities based on autonomy and coordination characteristics, building upon and refining existing taxonomies; (2) A catalogue of an initial set of design patterns for agentic AI systems that forms the basis of a new design methodology; (3) A clinical trial matching case study demonstrating pattern composition across all three tiers and (4) Formal grounding of Agentic Communities in ODP-EL communities, providing rigorous semantics for multi-participant coordination with deontic governance.

%
%

\section{Design Pattern Catalogue}
\label{sec:catalogue}

This section first introduces design pattern fundamentals we used in developing our design pattern catalogue and then provides an overview of the design patterns we developed so far. There are currently 46 patterns identified and complete documentation for all patterns is provided in the accompanying technical report.\footnote{The complete pattern catalogue with detailed mechanisms, 
compositional relationships, and implementation guidance is available as 
supplementary material accompanying this paper.} Representative patterns are discussed within this paper (Section~\ref{subsec:pattern-fundamentals} and Section~\ref{sec:demonstration}).

\subsection{Design Pattern Fundamentals}
\label{subsec:pattern-fundamentals}
\label{subsec:pattern-structure}

Design patterns capture recurring solutions to common problems within specific contexts \cite{gamma1995design}, describing: (1) the problem and context, (2) the forces that must be balanced, (3) the solution structure, and (4) the consequences of applying the pattern. Patterns work because they embody proven practice---solutions that have been discovered, applied, refined, and validated across multiple contexts, providing a level of abstraction above code that facilitates architectural reasoning and communication.

Each pattern in our catalogue follows a consistent documentation template that addresses both practical implementation and theoretical foundations. The template comprises six core elements:

\begin{itemize}[leftmargin=*]
\item \textit{Number}: A pattern identification number
\item \textit{What}: Core mechanism and purpose describing the pattern's essential functionality
\item \textit{Why}: Problem addressed and forces balanced, explaining when and why to use the pattern
\item \textit{Mechanisms}: Technology-agnostic implementation approach providing concrete guidance without prescribing specific tools
\item \textit{Agent Type Rationale}: Classification as LLM Agent, Agentic AI, or Agentic Community with detailed justification based on autonomy, learning, and coordination characteristics (reflecting the core characteristics from Table \ref{tab:three-tier-characteristics})
\item \textit{Related Patterns}: Identifies patterns that complement, require, or conflict with this pattern, supporting compositional relationships
\item \textit{References}: Grounds patterns in research literature and proven practice
\end{itemize}

All 46 patterns in the catalogue are documented using this template structure, ensuring consistency in documentation quality and facilitating pattern selection and composition. To illustrate this template structure within this paper, we present Pattern~\#1 (ReAct) as a representative example:

\textbf{Pattern \#1: ReAct (Reasoning and Acting)}

\textit{Agent Type}: Agentic AI

\textit{What}: A pattern that interleaves reasoning traces with action execution, allowing the agent to think through problems step-by-step while taking actions and observing their outcomes. The agent alternates between generating thoughts (reasoning about the current state and next action) and executing actions in the environment.

\textit{Why}: Enables more interpretable and robust decision-making by making the agent's reasoning process explicit. By reasoning before each action and observing results, agents can adapt their strategies dynamically rather than following rigid plans. The explicit reasoning traces provide transparency essential for debugging and trust in autonomous systems.

\textit{Mechanisms}: The agent follows thought-action-observation cycles where the language model generates reasoning traces explaining its decision before each action. Actions can include tool use, information retrieval, or environment manipulation. Observations from action results inform subsequent reasoning steps, creating a feedback loop that enables dynamic adaptation to unexpected outcomes.

\textit{Agent Type Rationale}: ReAct embodies Agentic AI through its combination of autonomous reasoning and adaptive action-taking. The pattern demonstrates genuine agency by independently determining which actions to take based on observations and dynamically adjusting its approach based on results. The iterative reasoning process shows goal-directed behavior with contextual decision-making---hallmarks of true agency beyond simple task execution.

\textit{Related Patterns}: Uses Tool-Using Agent (\#2) for action execution; enhanced by Memory-Augmented (\#3) for learning from past reasoning traces; forms foundation for Hierarchical Planning (\#5); used within Orchestration (\#14) for coordinated workflows; complements Reflexion (\#6) for self-improvement and Explanation (\#22) for reasoning transparency.

\textit{References}: Yao et al. (2022) "ReAct: Synergizing Reasoning and Acting in Language Models"

This structure ensures patterns are technology-agnostic (applicable across LLM providers and implementation languages), validated through research or practice, clearly scoped with well-defined relationships, and accessible to both researchers and practitioners.

\subsection{Catalogue Overview}
\label{subsec:catalogue-overview}

The patterns are developed through extensive analysis of recent AI developments and leveraging our experience in enterprise distributed systems standards, formal methods, interoperability frameworks, and industry practice. They embody the conceptual framework and formal 
semantics introduced in Section 2, providing practitioners with 
actionable architectural guidance grounded in rigorous foundations.
The complete list of all 46 patterns is presented in Table \ref{tab:pattern-catalogue}.

Detailed pattern descriptions with foundational references appear in the 
accompanying technical report; key patterns are discussed in 
Section~\ref{subsec:pattern-fundamentals} with representative citations 
grounding the catalogue in established research.

{\footnotesize  
\setlength{\tabcolsep}{3pt}  
\renewcommand{\arraystretch}{0.9}  
\begin{table}
\caption{Pattern Catalogue}
\label{tab:pattern-catalogue}
\begin{tabular}{@{}rlp{4.2cm}l@{}}  
\toprule
{\scriptsize\textbf{\#}} & {\scriptsize\textbf{Pattern}} & {\scriptsize\textbf{Purpose}} & {\scriptsize\textbf{Type}} \\
\midrule
\multicolumn{4}{@{}l}{\textit{Core Reasoning \& Learning (8 patterns)}} \\
1 & ReAct & Reasoning with actions & Agentic AI \\
3 & Memory-Augmented & Learning from interactions & Agentic AI \\
5 & Hierarchical Planning & Complex goal decomposition & Agentic AI \\
6 & Reflexion & Self-improvement & Agentic AI \\
7 & Constitutional AI & Value-aligned behavior & Agentic AI \\
8 & Critic-Actor & Generation + evaluation & Agentic AI \\
11 & Metacognitive & Reasoning about reasoning & Agentic AI \\
45 & Hybrid Neuro-Symbolic & Neural + symbolic reasoning & Agentic AI \\
\midrule
\multicolumn{4}{@{}l}{\textit{Tool \& Environment (2 patterns)}} \\
2 & Tool-Using & External tool access & LLM Agent \\
10 & Embodied & Environment interaction & Agentic AI \\
\midrule
\multicolumn{4}{@{}l}{\textit{Planning \& Decomposition (2 patterns)}} \\
37 & Decomposition & Task breakdown & Agentic AI \\
41 & Plan-then-Execute & Separate plan from execution & Agentic AI \\
\midrule
\multicolumn{4}{@{}l}{\textit{Coordination Mechanisms (6 patterns)}} \\
4 & Multi-Agent System & Agent coordination & Agentic Community \\
14 & Orchestration & Workflow management & Agentic Community \\
21 & Negotiation & Multi-party agreement & Agentic Community \\
35 & Ensemble & Combined robustness & Agentic Community \\
42 & Blackboard & Shared memory coordination & Agentic Community \\
43 & Debate/Deliberation & Structured debate & Agentic Community \\
\midrule
\multicolumn{4}{@{}l}{\textit{Communication Protocols (3 patterns)}} \\
12 & Inter-Agent Comm. & Agent interaction & Agentic Community \\
13 & Human-Agent Comm. & Human interaction & Agentic AI \\
15 & Semantic Bridge & Terminology translation & Agentic Community \\
\midrule
\multicolumn{4}{@{}l}{\textit{Governance \& Control (5 patterns)}} \\
18 & Compliance/Governance & Regulatory compliance & Agentic Community \\
19 & Access Control & Permission management & Agentic Community \\
20 & Audit Trail & Activity logging & Agentic Community \\
44 & Composable DSLs & Policy specification & Agentic Community \\
46 & Federated Privacy & Privacy-preserving training & Agentic Community \\
\midrule
\multicolumn{4}{@{}l}{\textit{Workflow Management (3 patterns)}} \\
9 & Workflow Agent & Structured processes & Agentic AI \\
33 & Batch Processing & Bulk processing & LLM Agent \\
34 & Real-Time/Streaming & Continuous streams & Agentic AI \\
\midrule
\multicolumn{4}{@{}l}{\textit{Quality \& Validation (2 patterns)}} \\
27 & Validation & Correctness verification & LLM Agent \\
31 & Error Recovery & Failure handling & Agentic AI \\
\midrule
\multicolumn{4}{@{}l}{\textit{Data Processing (4 patterns)}} \\
16 & Filtering/Triage & Information reduction & LLM Agent \\
17 & Structured Extraction & Unstructured to structured & LLM Agent \\
23 & Data Transformation & Format conversion & LLM Agent \\
26 & Summarization & Information condensation & LLM Agent \\
\midrule
\multicolumn{4}{@{}l}{\textit{Performance Optimization (4 patterns)}} \\
24 & Progressive Refinement & Iterative improvement & Agentic AI \\
32 & Fallback/Degradation & Graceful degradation & Agentic AI \\
36 & Caching/Memoization & Result reuse & LLM Agent \\
38 & Parallel Processing & Concurrent execution & LLM Agent \\
\midrule
\multicolumn{4}{@{}l}{\textit{Specialized Functions (4 patterns)}} \\
22 & Explanation & Human-understand. output & Agentic AI \\
25 & Version Control & Change tracking & LLM Agent \\
28 & Monitoring & System observability & LLM Agent \\
29 & Context Management & Context maintenance & Agentic AI \\
\midrule
\multicolumn{4}{@{}l}{\textit{Adaptation \& Learning (3 patterns)}} \\
30 & Adaptation & User customization & Agentic AI \\
39 & Feedback Loop & Feedback incorporation & Agentic AI \\
40 & Human-in-the-Loop & Human judgment integration & Agentic AI \\
\bottomrule
\end{tabular}
\end{table}
}


Analyzing the 46 patterns in Table \ref{tab:pattern-catalogue}, we observe the following distribution: 11 patterns (24\%) are of LLM Agent type; 22 patterns (48\%) are of Agentic AI type; and 13 patterns (28\%) are of Agentic Community type. This distribution suggests that nearly half of identified patterns represent genuine individual agentic AI with autonomous reasoning capabilities, while multi-participant coordination accounts for over a quarter---reflecting the reality that production systems increasingly require sophisticated community orchestration.

Agentic Community patterns coordinate heterogeneous participants including LLM Agents, Agentic AI entities, and human actors. In enterprise environments, humans function as domain experts, decision approvers, exception handlers, and supervisors within coordinated workflows---but also ultimate responsibility parties. Patterns like Negotiation, Audit Trail, and Compliance/Governance explicitly accommodate human participation, ensuring governance spans all participants.

\textbf{Accessing Pattern Details:} The patterns listed in 
Table~\ref{tab:pattern-catalogue} are fully documented in supplementary 
material using the template structure described in 
Section~\ref{subsec:pattern-fundamentals}. Each pattern includes detailed 
implementation mechanisms, agent type classification rationale grounded in 
the three-tier framework, compositional relationships with related patterns, 
and references to foundational research. Representative patterns are discussed 
within this paper: Pattern~\#1 (ReAct) illustrates the template in 
Section~\ref{subsec:pattern-fundamentals}, while Section~\ref{sec:demonstration} 
demonstrates pattern composition across all three tiers in the clinical trial 
matching case study. Complete documentation for all 46 patterns, including 
comprehensive references, mechanisms, and usage guidance, is available in 
supplementary material.

\subsection{Pattern Categories and Organization}
\label{subsec:categories}

Table \ref{tab:pattern-catalogue} also shows that our 46 patterns are organised into 12 thematic categories that address distinct architectural concerns. These categories group into four high-level areas reflecting key architectural themes, as shown in Table \ref{tab:pattern-categories}. Each of the 12 themes addresses specific architectural concerns:

\textbf{Foundational Cognitive Patterns} provide the cognitive foundation for agentic behavior:

\textit{Core Reasoning \& Learning}: Patterns enabling agents to think, reason, learn from experience, and improve through self-reflection. Characterized by internal cognitive capabilities.

\textit{Tool \& Environment}: Patterns enabling agents to interact with external tools, APIs, and physical/simulated environments. Characterized by external capability extension.

\textit{Planning \& Decomposition}: Patterns for breaking complex, long-horizon goals into manageable subgoals and execution plans. Characterized by hierarchical goal structures.

\textbf{Multi-Agent Coordination Patterns} enable systems to scale beyond single agents:

\textit{Coordination Mechanisms}: Patterns for orchestrating multiple specialized agents to achieve collective goals. Characterized by system-level coordination mechanisms.

\textit{Communication Protocols}: Patterns for structured interaction between agents or with humans. Characterized by formal message exchange and protocol adherence.

\textbf{Governance and Safety Patterns} ensure compliant, secure, and auditable behavior:

\textit{Governance \& Control}: Patterns for policy enforcement, access control, audit trails, and regulatory compliance. Characterized by formal governance frameworks using ODP-EL deontic tokens.

\textbf{Specialized Functional Patterns} address operational requirements:

\textit{Workflow Management}: Patterns for executing structured, multi-step processes with state management. Characterized by explicit process definitions and execution tracking.

\textit{Quality \& Validation}: Patterns for ensuring correctness, detecting errors, and maintaining quality standards. Characterized by verification and assurance mechanisms.

\textit{Data Processing}: Patterns for transforming, filtering, extracting, and validating data across formats. Characterized by data pipeline operations.

\textit{Performance Optimization}: Patterns for improving throughput, reducing latency, and efficient resource utilization. Characterized by computational efficiency techniques.

\textit{Specialized Functions}: Patterns for specific capabilities like explanation, versioning, monitoring, and context management. Characterized by targeted functional utilities.

\textit{Adaptation \& Learning}: Patterns for continuous improvement, personalization, and feedback incorporation. Characterized by system evolution over time.

\begin{table}[t]
\centering
\caption{Pattern Categories: Hierarchical Organization}
\label{tab:pattern-categories}
\small
\begin{tabular}{@{}lp{4cm}ll@{}}  
\toprule
\multicolumn{1}{c}{\textbf{Group (Count)}} & \multicolumn{1}{c}{\textbf{Category}} & \multicolumn{1}{c}{\textbf{Purpose}} & \multicolumn{1}{c}{\textbf{Patterns}} \\
\midrule
\multirow{3}{*}{\begin{tabular}[c]{@{}l@{}}\textbf{Foundational}\\[-0.5ex]\textbf{Cognitive (12)}\end{tabular}} 
& Core Reasoning \& Learning & Fundamental reasoning & 8 \\
& Tool \& Environment & External interaction & 2 \\
& Planning \& Decomposition & Goal decomposition & 2 \\
\midrule
\multirow{2}{*}{\begin{tabular}[c]{@{}l@{}}\textbf{Multi-Agent}\\[-0.5ex]\textbf{Coordination (9)}\end{tabular}} 
& Coordination Mechanisms & Multi-agent collaboration & 6 \\
& Communication Protocols & Structured communication & 3 \\
\midrule
{\begin{tabular}[c]{@{}l@{}}\textbf{Governance}\\[-0.5ex]\textbf{\& Safety (5)}\end{tabular}} 
& Governance \& Control & Policy enforcement & 5 \\
\midrule
\multirow{6}{*}{\begin{tabular}[c]{@{}l@{}}\textbf{Specialized}\\[-0.5ex]\textbf{Functional (20)}\end{tabular}} 
& Workflow Management & Process management & 3 \\
& Quality \& Validation & Correctness assurance & 2 \\
& Data Processing & Data handling & 4 \\
& Performance Optimization & Efficiency & 4 \\
& Specialized Functions & Specific capabilities & 4 \\
& Adaptation \& Learning & Continuous improvement & 3 \\

\bottomrule
\end{tabular}
\end{table}

\section{Agentic AI Architecture Design Methodology}
\label{sec:design-methodology}

\subsection{Overview}

\begin{figure}
    \centering
    \includegraphics[width=0.75\linewidth]{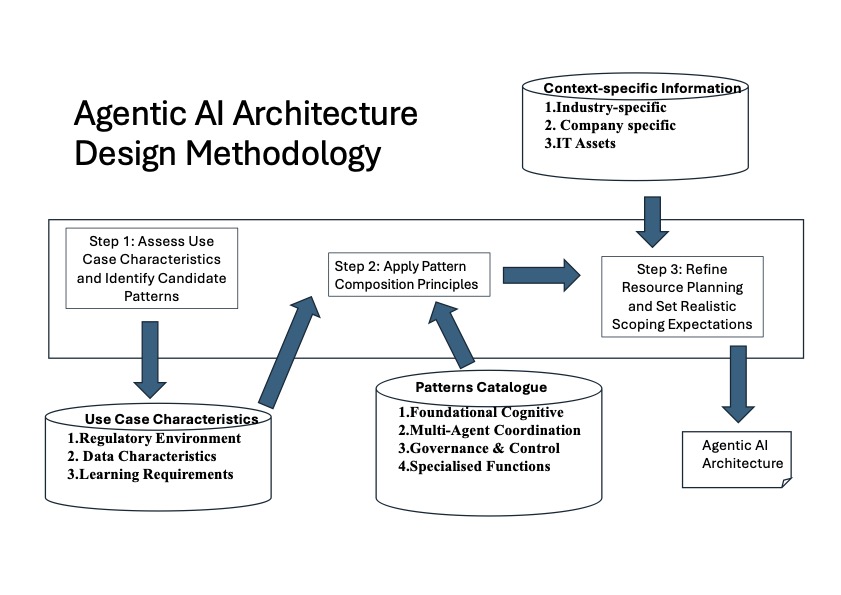}
    \caption{Applying Design Patterns for Producing an Agentic AI Architecture}
    \label{fig:Methodology}
\end{figure}

We now present a general methodology for applying design patterns to architecting enterprise agentic AI solutions. This methodology involves a three-step approach---assess characteristics, compose patterns, and apply appropriate scoping---providing systematic guidance for pattern selection and composition applicable to any domain. Figure \ref{fig:Methodology} illustrates the overall process.

\subsection{Step 1: Assess Use Case Characteristics and Identify Candidate Patterns}
\label{subsec:assess-characteristics}

Pattern selection begins by systematically evaluating four key dimensions that determine architectural requirements. Each dimension maps to specific pattern categories, enabling architects to identify relevant patterns based on use case properties:

\begin{enumerate}
    \item 
\textbf{Autonomy Requirements} determine whether tasks require simple execution, strategic reasoning, or coordinated multi-agent behavior. Low autonomy tasks involving well-defined data processing with clear input-output mappings require LLM Agent patterns such as Structured Extraction (\#17), Data Transformation(\#23), and Validation (\#27). These patterns provide reliable task execution without requiring autonomous decision-making. High autonomy requirements involving strategic decisions, complex reasoning, or adaptive behavior necessitate Agentic AI patterns such as ReAct (\#1) for reasoning-action cycles, Hierarchical Planning (\#5) for goal decomposition, and Memory-Augmented (\#3) for learning from experience. When problems require specialized agents coordinating toward shared goals, Agentic Community patterns become essential, including Multi-Agent System (\#4) for overall coordination, Orchestration (\#14) for workflow management, and Inter-Agent Communication (\#12) for structured interaction.
\item
\textbf{Regulatory Environment} dictates governance requirements that must be embedded throughout the architecture. Regulated industries including healthcare (HIPAA), finance (SEC/FINRA), and legal domains mandate a comprehensive governance cluster spanning Constitutional AI (\#7) for value alignment, Compliance/Governance (\#18) for policy enforcement, Access Control (\#19) for authorization management, and Audit Trail (\#20) for complete traceability. High-risk decisions involving patient safety, financial transactions, or legal consequences require additional safeguards through Validation (\#27) for correctness verification and Human-in-the-Loop (\#40) ensuring human authority over critical decisions. Explanation (\#22) become mandatory in regulated contexts where decision transparency and justification are required for compliance documentation.
\item
\textbf{Data Characteristics} determine processing pipeline requirements based on data structure, volume, and latency constraints. Structured data processing employs data pipeline patterns including Filtering/Triage (\#16) for early volume reduction, Structured Extraction (\#17) for format conversion, Data Transformation (\#23) for schema mapping, and Validation (\#27) for quality assurance. Mixed structured and unstructured data requires Hybrid Neuro-Symbolic (\#45) combining statistical learning with symbolic reasoning. High-volume batch processing benefits from Batch Processing (\#33) and Parallel Processing (\#38) patterns, while real-time interaction demands Real-Time/Streaming (\#34) for responsiveness. Data diversity requiring cross-domain integration necessitates Semantic Bridge (\#15) for terminology translation.
\item
\textbf{Learning Requirements} specify whether systems remain static or evolve through experience. Static behavior suffices for well-defined tasks using basic patterns such as ReAct (\#1) for reasoning and Tool-Using (\#2) for capability extension. Continuous improvement from user feedback or outcome observation requires learning patterns including Memory-Augmented (\#3) for experience retention, Reflexion (\#6) for self-critique and improvement, and Feedback Loop (\#39) for systematic incorporation of corrections. Adaptive personalization tailoring behavior to individual users employs Adaptation (\#30) combined with Context Management (\#29) for maintaining user-specific preferences across interactions. Systems requiring ongoing refinement benefit from Progressive Refinement (\#24) enabling iterative quality improvement.
\end{enumerate}

This assessment produced a candidate set of patterns addressing the use case's 
autonomy, regulatory, data, and learning requirements. Step 2 now composes 
these candidates into a coherent architecture.

\subsection{Step 2: Apply Pattern Composition Principles}
\label{subsec:compose-patterns}

Individual patterns rarely suffice in isolation. Having identified candidate 
patterns through characteristic assessment in Step 1, patterns must be composed 
into coherent architectures that ensure completeness, compatibility, and 
appropriate governance. Understanding how patterns relate to each other is 
fundamental to effective composition.

Patterns interact through three fundamental relationship types that guide 
composition decisions:

\begin{itemize}[leftmargin=*]
\item \textbf{Layered relationships}: One pattern builds upon and incorporates 
another's capabilities as a foundation. The dependent pattern extends the base 
pattern with additional functionality. Example: ReAct (\#1) builds on Tool-Using 
(\#2) by adding reasoning traces and iterative action-observation cycles to tool 
calling capabilities. ReAct systems inherently include tool-using functionality 
enhanced with explicit reasoning.

\item \textbf{Complementary relationships}: Patterns work together in parallel, 
each addressing different aspects of system behavior to create combined benefits 
neither achieves alone. Example: Memory-Augmented (\#3) combined with 
Constitutional AI (\#7) creates agents that learn from experience while 
maintaining value alignment---the memory enables improvement while constitutional 
principles ensure learned behaviors remain ethically bounded.

\item \textbf{Alternative relationships}: Patterns address the same architectural 
concern through different approaches, requiring selection of one based on use 
case constraints. Example: Batch Processing (\#33) versus Real-Time/Streaming 
(\#34) both handle data processing workloads but make different latency-throughput 
trade-offs. Systems typically choose one based on whether batch or real-time 
processing better matches requirements.
\end{itemize}

These relationship types inform three primary composition strategies:

\label{subsec:composition-evolution}
The first strategy employs three composition techniques that leverage the pattern 
relationship types described above:

\begin{itemize}
\item \textbf{Vertical Composition}---layering patterns from simple to complex 
through layered relationships, building sophisticated capabilities atop 
foundational ones (Example: Tool-Using Agent $\rightarrow$ ReAct $\rightarrow$ 
Multi-Agent Community);

\item \textbf{Horizontal Composition}---combining peer patterns through 
complementary relationships for additive capabilities within the same tier 
(Example: Memory-Augmented + Constitutional AI creates value-aligned agents that 
learn from experience within ethical boundaries);

\item \textbf{Cross-Cutting Composition}---overlaying governance patterns across 
functional patterns to ensure system-wide compliance (Example: Audit Trail + 
Compliance/Governance patterns spanning all agents in a community).
\end{itemize}

The second strategy follows evolutionary development paths that progressively add 
sophistication: (1) Start with narrow LLM Agents for specific, well-scoped tasks 
(data validation, extraction, transformation); (2) Add agentic capabilities 
(planning, reasoning, learning) as requirements grow and single-task automation 
proves insufficient, introducing ReAct, Memory-Augmented, and Hierarchical 
Planning patterns; (3) Scale through Agentic Community coordination when single 
agents prove insufficient for complex problems, introducing Orchestration, 
Multi-Agent System, and Inter-Agent Communication patterns; and (4) Overlay 
governance patterns as compliance and accountability needs emerge with system 
maturity, adding Constitutional AI, Compliance/Governance, Access Control, and 
Audit Trail patterns. This evolutionary approach reduces risk by validating 
simpler patterns before introducing coordination complexity.

The third strategy (applicable for complex systems requiring cross-organizational 
integration) applies a three-layer reference architecture that reflects our 
pattern categorization:

\begin{enumerate}
\item \textbf{Foundation Layer}: manages standardized data access and security 
boundaries, comprising LLM Agent patterns for data processing (Extraction, 
Transformation, Validation) and essential governance (Access Control, basic 
Compliance/Governance). It provides technical interoperability with external 
systems.

\item \textbf{Core Processing Layer}: manages autonomous reasoning and 
decision-making, intra-organizational coordination and essential governance 
(Constitutional AI, Human-in-the-Loop for critical decisions). It also supports 
coordination, orchestration and inter-agent communication.

\item \textbf{External Integration Layer}: manages dynamic external coordination 
using advanced Agentic Community patterns (Negotiation, Semantic Bridge). It is 
responsible for cross-organizational interaction and dynamic capability discovery.
\end{enumerate}

Depending on use-case characteristics (Step 1), there can be variations. For 
example, simple systems may collapse Layers 2-3 or operate with only Layers 1-2, 
internal-only systems may not require Layer 3, and systems without structured 
data may not need Layer 1.

Pattern composition produces a comprehensive architecture specification. 
Step 3 determines how to scope this architecture for incremental delivery 
based on available resources and risk tolerance.

\subsection{Step 3: Apply Scale-Appropriate Scoping and Implementation Planning}
\label{subsec:apply-scoping}

Step 2 usually produces a comprehensive architecture that cannot be implemented in one go. Pattern scoping addresses three interconnected decisions: determining system complexity tier based on use case characteristics, prioritizing patterns within that tier for maximum value delivery, and sequencing implementation across development phases to manage risk and prove value incrementally.

Scoping should first identify a {\it complexity tier} based on use case characteristics to avoid both under-architecting (insufficient capabilities) and over-architecting (unnecessary complexity). Based on our analysis of production systems and industry practice, we provide scoping guidance for three complexity tiers.

\textbf{Simple Automation} (3--5 patterns) addresses well-defined, narrow tasks with limited autonomy requirements. Typical deployments implement the essential core subset: one reasoning pattern (ReAct or Tool-Using), Validation for quality assurance, and optionally Memory-Augmented for context retention. Examples include document processing pipelines, data extraction services, and automated classification tasks. These systems prove their value quickly with minimal architectural complexity while establishing foundations for future enhancement.

\textbf{Departmental Applications} (8--12 patterns) serve organizational units requiring moderate autonomy and coordination. Beyond the essential core, these systems add multi-agent coordination (Multi-Agent System, Inter-Agent Communication, Orchestration), learning capabilities (Reflexion, Feedback Loop), and specialized functions (Explanation, Context Management). Examples include customer service assistants, internal knowledge management systems, and workflow automation platforms. These deployments balance sophistication with maintainability, supporting teams without requiring extensive AI operations infrastructure.

\textbf{Enterprise-Wide Systems} (15--20 patterns) in regulated industries require comprehensive pattern composition spanning all categories. Full governance clusters (Constitutional AI, Compliance/Governance, Access Control, Audit Trail, Composable DSLs) combine with sophisticated coordination mechanisms (Orchestration, Negotiation, Semantic Bridge), robust error handling (Error Recovery, Fallback/Graceful Degradation, Ensemble), and continuous learning (Memory-Augmented, Reflexion, Adaptation/Personalization, Human-in-the-Loop). Examples include clinical decision support systems, algorithmic trading platforms, and multi-agent manufacturing control systems. These deployments justify complexity through business-critical functionality and regulatory requirements.

Table \ref{tab:scoping-guidance} summarizes pattern counts and characteristics across complexity tiers, providing planning guidance for resource allocation and iterative delivery. It also shows that each complexity tier is prioritized based on business value, risk mitigation, and implementation dependencies using Essential (e.g. for MVP), Important (e.g. for production) and Optimizing (e.g. for long-term maintenance). This prioritization enables development teams to focus initial development on value-delivering capabilities while deferring optimization patterns until production experience justifies their complexity.

\begin{table}[t]
\centering
\caption{Pattern Composition Scoping Guidance}
\label{tab:scoping-guidance}
\small
\begin{tabular}{@{}llll@{}}
\toprule
\textbf{Tier} & \textbf{Pattern Count} & \textbf{Key Additions} & \textbf{Priority} \\
\midrule
Simple Automation & 3--5 
& + Basic reasoning & Essential \\
& & + Quality Assurance & (must have for MVP) \\
& & + Context Retention & Small team (2--3 developers)\\
\midrule
Departmental & 8--12 & 
+ Multi-agent coord. & Important \\
& & + Learning capabilities & (add for production readiness) \\
& & + Specialised functions & Medium team (4--6 developers)\\
\midrule
Enterprise-Wide & 15--20 & 
+ Full governance & Optimising \\
(Regulated) & & + Advanced coord. & (enhance over time) \\
& & + Error handling &  Large team (6--10 developers)\\
& & + Continuous learning &  \\
\bottomrule
\end{tabular}
\end{table}

The clinical trial matching system described in Section \ref{sec:demonstration} exemplifies different tier complexity with 15 patterns spanning governance, multi-agent coordination, and specialized functions required for regulated healthcare deployment. Tier 1 established FHIR data access and basic matching; Tier 2 added governance, orchestration, and physician integration; Tier 3 introduced external negotiation and continuous learning capabilities.

\section{Demonstration: Clinical Trial Matching Community}
\label{sec:demonstration}

This section validates our pattern-based design methodology through a clinical 
trial matching system. We demonstrate how the three-step methodology from 
Section \ref{sec:design-methodology} guided architectural decisions from 
requirements analysis through implementation. Section \ref{subsec:usecase-overview} 
presents the healthcare use case, while the following three subsections apply 
each methodology step sequentially. This demonstrates patterns functioning as 
genuine design tools that shape architectural choices, not merely as post-hoc 
labels for completed systems.

\subsection{Use Case Overview}
\label{subsec:usecase-overview}

Clinical trial recruitment faces significant challenges: trials fail to meet enrollment targets 80\% of the time, while many eligible patients remain unaware of relevant studies \cite{fogel2018factors}. Manual matching of patient records against trial eligibility criteria is time-consuming, error-prone, and doesn't scale.

When combined with structural approach for representing patient information, such as the one based on the HL7 FHIR standards, an agentic AI approach offers compelling advantages. It allows autonomous reasoning over complex medical criteria expressed in natural text, dynamic adaptation to new trial protocols, integration across heterogeneous healthcare systems (EHRs, trial registries, laboratory systems), seamless coordination with healthcare professionals, and compliance with healthcare regulations (HIPAA, informed consent).

We now demonstrate how the three-step methodology from Section \ref{sec:design-methodology} guided our architecture design for this use case.

\subsection{Step 1: Use Case Characteristic Assessment}

\textbf{Autonomy Requirements Analysis:}
\begin{itemize}[leftmargin=*]
\item \textit{Low autonomy tasks} (LLM Agent patterns): Data extraction from FHIR resources requires Structured Extraction (\#17); format transformation requires Data Transformation (\#23); and data quality assurance requires Validation (\#27)---these provide reliable execution within well-defined parameters without strategic decision-making.

\item \textit{High autonomy tasks} (Agentic AI patterns): Understanding complex eligibility criteria and reasoning about patient-trial compatibility requires ReAct (\#1) for reasoning-action cycles; decomposing matching tasks into manageable subgoals requires Hierarchical Planning (\#5); and maintaining context across patient evaluations requires Memory-Augmented (\#3) for experience retention---these require genuine autonomous judgment.

\item \textit{Multi-agent coordination} (Agentic Community patterns): Coordinating specialized agents requires Orchestration (\#14) for workflow management; structured interaction between agents requires Inter-Agent Communication (\#12); dynamic coordination with external trial sites requires Negotiation (\#21); integrating physician oversight at critical decision points requires Human-in-the-Loop (\#40)---these require system-level orchestration with formal governance.
\end{itemize}

\textbf{Regulatory Environment Analysis:}
As a healthcare application subject to HIPAA regulations with potential patient safety implications, the system requires:
\begin{itemize}[leftmargin=*]
\item \textit{Mandatory governance cluster}: Compliance/Governance (\#18) for HIPAA enforcement; Access Control (\#19) via SMART on FHIR for authorization management; Audit Trail (\#20) for complete traceability and accountability

\item \textit{High-risk decision safeguards}: Validation (\#27) for eligibility assessment correctness; Human-in-the-Loop (\#40) ensuring physicians make final enrollment decisions; Explanation (\#22) for clinical transparency and decision justification
\end{itemize}

\textbf{Data Characteristics Analysis:}
\begin{itemize}[leftmargin=*]
\item \textit{Structured data processing}: Patient demographics, lab results, vital signs in FHIR format require Structured Extraction (\#17) for format conversion; Data Transformation (\#23) for schema mapping; and Validation (\#27) for quality assurance

\item \textit{Unstructured data processing}: Clinical notes and eligibility criteria text require ReAct (\#1) for reasoning-action cycles; and Semantic Bridge (\#15) for terminology translation across clinical, research, and administrative domains

\item \textit{Multi-domain integration}: Coordinating across heterogeneous healthcare systems requires Semantic Bridge (\#15) for harmonizing terminologies; and Inter-Agent Communication (\#12) for structured interaction protocols
\end{itemize}

\textbf{Learning Requirements Analysis:}
\begin{itemize}[leftmargin=*]
\item \textit{Context retention}: Maintaining relevant information across patient evaluations and trial assessments requires Memory-Augmented (\#3) for experience retention and context management

\item \textit{Decision transparency}: Providing clinically meaningful explanations requires Explanation (\#22) for translating agent reasoning into physician-appropriate justifications
\end{itemize}

This assessment revealed requirements spanning all three tiers: LLM Agents for data processing, Agentic AI for medical reasoning, and Agentic Communities for multi-participant coordination with governance.

\subsection{Step 2: Pattern Composition Design}
\label{subsec:step2-composition} 

Based on the characteristic assessment, we composed patterns across three architectural layers:

\textbf{Layer 1: FHIR Foundation (Data Standards + Governance)}
\begin{itemize}[leftmargin=*]
\item \textit{Data processing patterns}: Structured Extraction (\#17) for FHIR resource extraction; Data Transformation (\#23) for format conversion
\item \textit{Governance and control patterns}: Access Control (\#19) via SMART on FHIR OAuth2; Compliance/Governance (\#18) via Consent Resource management; Audit Trail (\#20) for data access logging
\item \textit{Rationale}: Isolates healthcare interoperability and regulatory compliance complexities, ensuring all data access respects patient consent and authorization boundaries with complete traceability
\end{itemize}

\textbf{Layer 2: Matching Workflow (Agentic AI Core + Coordination)}
\begin{itemize}[leftmargin=*]
\item \textit{Core reasoning patterns}: ReAct (\#1) for reasoning-action cycles in eligibility assessment; Memory-Augmented (\#3) for context retention across evaluations; Hierarchical Planning (\#5) for decomposing complex matching tasks
\item \textit{Quality and transparency}: Validation (\#27) for correctness verification; Explanation (\#22) for clinical transparency
\item \textit{Cross-domain harmonization}: Semantic Bridge (\#15) for terminology translation between clinical and research vocabularies
\item \textit{Human integration}: Human-in-the-Loop (\#40) ensures physicians retain final enrollment authority
\item \textit{Rationale}: Vertical composition builds from data extraction (LLM Agents) through autonomous reasoning (Agentic AI) to coordinated workflow (Agentic Community), demonstrating the full three-tier framework
\end{itemize}

\textbf{Layer 3: Conversational Negotiation (External Integration)}
\begin{itemize}[leftmargin=*]
\item \textit{Dynamic coordination}: Negotiation (\#21) for dynamic coordination with trial sites and external systems; Inter-Agent Communication (\#12) for structured interaction protocols; Orchestration (\#14) for managing external workflows
\item \textit{Semantic interoperability}: Semantic Bridge (\#15) for terminology translation across clinical, research, and administrative domains
\item \textit{Governance enforcement}: Compliance/Governance (\#18) spanning all external interactions; Audit Trail (\#20) for complete provenance; Access Control (\#19) for authorization boundaries
\item \textit{Rationale}: Demonstrates sophisticated Agentic Community patterns addressing real-world enterprise integration challenges---heterogeneous systems, cross-organizational boundaries, regulatory constraints
\end{itemize}

These three layers represent our architectural design decisions based on 
pattern composition principles. Subsection \ref{subsec:three-layer-architecture} 
formalizes this architecture by specifying each layer as an ODP-EL community 
with explicit roles, normative constraints (deontic tokens), contracts, and 
community objects---transforming the conceptual design into verifiable formal 
specifications.

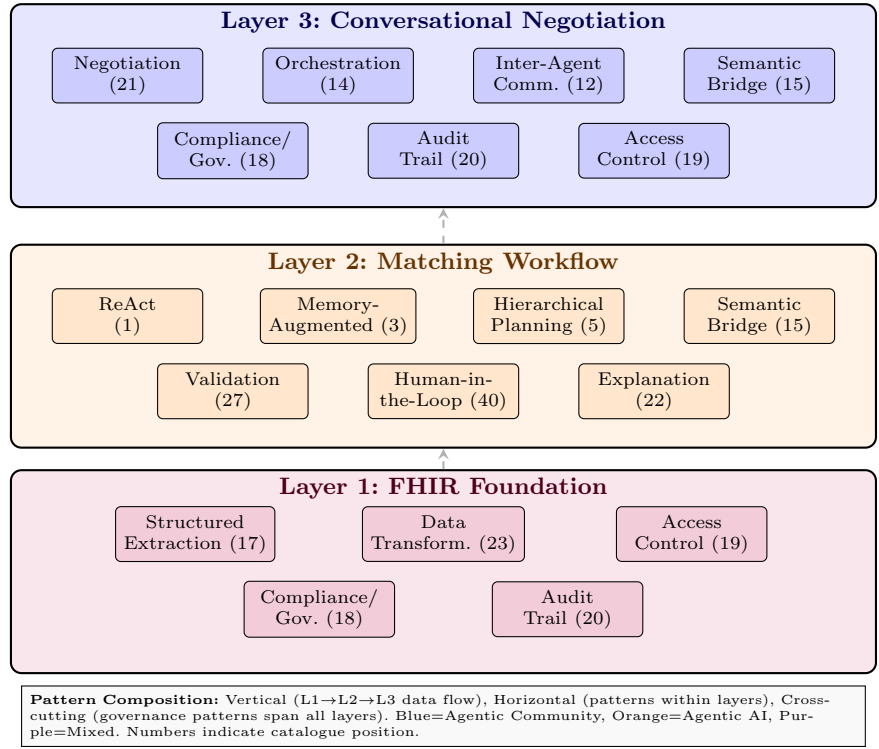
\begin{figure}[t]
\centering
\begin{tikzpicture}[
  layer/.style={rectangle, draw, thick, rounded corners, minimum width=11.5cm, minimum height=2.7cm},
  pattern/.style={rectangle, draw, fill=white, rounded corners=2pt, minimum width=2cm, minimum height=0.58cm, align=center, font=\scriptsize},
  arrow/.style={->, >=stealth, thick}
]

\node[layer, fill=blue!10] (l3) at (0,5.5) {};
\node[below=0.0cm, font=\small\bfseries, text=blue!30!black] at (l3.north) {Layer 3: Conversational Negotiation};

\node[pattern, fill=blue!20] at (-4.2,5.9) {Negotiation\\(21)};
\node[pattern, fill=blue!20] at (-1.4,5.9) {Orchestration\\(14)};
\node[pattern, fill=blue!20] at (1.4,5.9) {Inter-Agent\\Comm. (12)};
\node[pattern, fill=blue!20] at (4.2,5.9) {Semantic\\Bridge (15)};

\node[pattern, fill=blue!20] at (-2.8,4.9) {Compliance/\\Gov. (18)};
\node[pattern, fill=blue!20] at (0,4.9) {Audit\\Trail (20)};
\node[pattern, fill=blue!20] at (2.8,4.9) {Access\\Control (19)};

\node[layer, fill=orange!10] (l2) at (0,2.3) {};
\node[below=0.0cm, font=\small\bfseries, text=orange!40!black] at (l2.north) {Layer 2: Matching Workflow};

\node[pattern, fill=orange!20] at (-4.2,2.7) {ReAct\\(1)};
\node[pattern, fill=orange!20] at (-1.4,2.7) {Memory-\\Augmented (3)};
\node[pattern, fill=orange!20] at (1.4,2.7) {Hierarchical\\Planning (5)};
\node[pattern, fill=orange!20] at (4.2,2.7) {Semantic\\Bridge (15)};

\node[pattern, fill=orange!20] at (-2.8,1.7) {Validation\\(27)};
\node[pattern, fill=orange!20] at (0,1.7) {Human-in-\\the-Loop (40)};
\node[pattern, fill=orange!20] at (2.8,1.7) {Explanation\\(22)};

\node[layer, fill=purple!10] (l1) at (0,-0.7) {};
\node[below=0.0cm, font=\small\bfseries, text=purple!40!black] at (l1.north) {Layer 1: FHIR Foundation};

\node[pattern, fill=purple!20] at (-3.3,-0.2) {Structured\\Extraction (17)};
\node[pattern, fill=purple!20] at (0,-0.2) {Data\\Transform. (23)};
\node[pattern, fill=purple!20] at (3.3,-0.2) {Access\\Control (19)};

\node[pattern, fill=purple!20] at (-1.65,-1.2) {Compliance/\\Gov. (18)};
\node[pattern, fill=purple!20] at (1.65,-1.2) {Audit\\Trail (20)};

\draw[arrow, dashed, gray!60] (l1.north) -- (l2.south);
\draw[arrow, dashed, gray!60] (l2.north) -- (l3.south);

\node[rectangle, draw, fill=gray!5, align=left, font=\tiny, text width=11cm, below=0.15cm of l1.south] {
\textbf{Pattern Composition:} Vertical (L1$\rightarrow$L2$\rightarrow$L3 data flow), Horizontal (patterns within layers), 
Cross-cutting (governance patterns span all layers). Blue=Agentic Community, Orange=Agentic AI, Purple=Mixed. 
Numbers indicate catalogue position.
};

\end{tikzpicture}
\caption{Pattern composition in clinical trial matching system demonstrating vertical composition (Layer 1 $\rightarrow$ 2 $\rightarrow$ 3), horizontal composition (patterns within layers), and cross-cutting composition (governance patterns spanning multiple layers).}
\label{fig:pattern-composition}
\end{figure}

In total, this healthcare use case required 15 patterns (shown in Figure \ref{fig:pattern-composition}) spanning all categories.

\subsection{Step 3: Pattern Scoping and Tier Allocation}

Applying the complexity tier framework from Section \ref{subsec:apply-scoping}, we allocated the 15 identified patterns across three implementation tiers to enable incremental delivery while managing risk. This allocation follows the principle of starting with essential capabilities that prove value quickly, then adding production-readiness features, and finally extending to full enterprise integration. Table \ref{tab:scoping-application} shows the tier allocation with rationale for each pattern's placement.

\begin{table}[t]
\centering
\caption{Pattern Tier Allocation for Clinical Trial Matching System}
\label{tab:scoping-application}
\small
\begin{tabular}{@{}p{2.2cm}p{1.2cm}p{4.5cm}p{5.5cm}@{}}
\toprule
\textbf{Tier} & \textbf{Count} & \textbf{Patterns} & \textbf{Rationale} \\
\midrule
\multirow{5}{2.2cm}{Tier 1:\\Simple Automation\\} & \multirow{5}{1.2cm}{5 patterns} 
& ReAct (\#1) & Core reasoning foundation for eligibility assessment \\
& & Structured Extraction (\#17) & FHIR data access essential for any matching \\
& & Data Transform. (\#23) & Format conversion between FHIR and matching engine \\
& & Validation (\#27) & Quality assurance prevents error propagation \\
& & Memory-Aug. (\#3) & Context retention across patient evaluations \\
\midrule
\multirow{6}{2.2cm}{Tier 2:\\Departmental\\Application\\} & \multirow{6}{1.2cm}{+5 patterns\\(total: 10)} 
& Access Control (\#19) & SMART on FHIR authorization for HIPAA compliance \\
& & Compliance/Gov. (\#18) & Consent management and regulatory enforcement \\
& & Audit Trail (\#20) & Complete traceability for accountability \\
& & Human-in-Loop (\#40) & Physician final authority for patient safety \\
& & Explanation (\#22) & Clinical transparency for decision justification \\
\midrule
\multirow{5}{2.2cm}{Tier 3:\\Enterprise-Wide\\} & \multirow{5}{1.2cm}{+5 patterns\\(total: 15)} 
& Orchestration (\#14) & Multi-agent workflow coordination \\
& & Inter-Agent Comm. (\#12) & Structured protocols between agents \\
& & Negotiation (\#21) & Dynamic coordination with external trial sites \\
& & Hierarchical Plan. (\#5) & Complex task decomposition for multi-trial scenarios \\
& & Semantic Bridge (\#15) & Terminology translation across domains \\
\bottomrule
\end{tabular}
\end{table}

\textbf{Tier 1 (Simple Automation)} establishes the minimum viable product demonstrating core value: matching patients to trials through FHIR data access, reasoning-based eligibility assessment, and quality validation. This 5-pattern configuration proves feasibility while minimizing architectural complexity, enabling rapid validation of the approach with clinical stakeholders. Memory-Augmented pattern inclusion ensures context retention across multiple patient evaluations within a session, critical for comparative assessment of trial options.

\textbf{Tier 2 (Departmental Application)} adds production readiness through comprehensive governance and human integration. The governance cluster (Access Control, Compliance/Governance, Audit Trail) ensures HIPAA compliance and creates accountability chains essential for healthcare deployment. Human-in-the-Loop and Explanation patterns integrate physician oversight at decision points while providing clinical transparency---addressing the reality that autonomous AI recommendations require human validation in patient care contexts. This 10-pattern configuration represents a production-ready system deployable within a single healthcare organization.

\textbf{Tier 3 (Enterprise-Wide System)} extends to multi-site deployment requiring sophisticated agent coordination and semantic interoperability. Orchestration and Inter-Agent Communication enable coordinated workflows across specialized agents handling patient screening, eligibility assessment, and external communications. Negotiation pattern facilitates dynamic coordination with external trial sites lacking standardized APIs, critical for real-world clinical trial networks. Hierarchical Planning manages complexity when matching patients against hundreds of concurrent trials, while Semantic Bridge resolves terminology differences between clinical practice, research protocols, and administrative systems. This 15-pattern configuration represents the full enterprise architecture demonstrated in Figure \ref{fig:pattern-composition}.

The tier allocation validates our scoping guidance (Table \ref{tab:scoping-guidance}) while demonstrating incremental value delivery. Note that governance patterns (Compliance/Governance, Access Control, Audit Trail) appear as cross-cutting concerns spanning Layers 1 and 3 in the architectural view (Figure \ref{fig:pattern-composition}), while Semantic Bridge spans Layers 2 and 3, demonstrating how implementation tiers differ from architectural layers---tiers reflect deployment phases while layers represent logical system organization.

\label{sec:formal-analysis}

\section{ODP-EL: Formal Foundation for Agentic Community Architecture}
\label{sec:formal-analysis}

\subsection{Introduction}
\label{sec:formal-analysis-intro}

Enterprise-grade agentic AI systems require rigorous formal foundations to ensure verifiable governance, accountability, and regulatory compliance. We ground our architectural approach in the ISO Open Distributed Processing Enterprise Language (ODP-EL) standard \cite{IS15414}---a proven, internationally standardized framework for specifying enterprise distributed systems with formal semantics.

ODP-EL provides the essential formal foundation that our pattern catalogue builds upon. While design patterns (Section \ref{sec:catalogue}) offer practical architectural guidance for common problems, ODP-EL provides the formal rigor necessary for enterprise deployment. This combination delivers both practicality and verifiability: patterns guide architectural decisions while ODP-EL specifications enable formal verification of governance properties, accountability chains, and compliance requirements.

The ODP-EL framework addresses critical requirements for enterprise agentic AI systems:

\textbf{Rigorous Architecture Specification}: ODP-EL communities provide precise formal semantics for specifying multi-participant coordination frameworks. Roles define behavioral specifications fillable by LLM Agents, Agentic AI, or human actors. Contracts bind roles through normative relationships, while policies express governance constraints applicable to all participants regardless of whether they are computational or human agents.

\textbf{Verifiable Governance Properties}: Deontic logic embedded in ODP-EL enables formal expression and verification of obligations (burdens), permissions (permits), and prohibitions (embargoes). These deontic constraints create traceable accountability chains essential for regulated domains, with formal semantics enabling automated verification that governance policies are enforced correctly.

\textbf{Standards-Based Design}: As an ISO standard, ODP-EL provides internationally recognized semantics for enterprise system specification. This standards foundation ensures our approach builds on proven enterprise architecture practice rather than ad-hoc formalism, facilitating regulatory acceptance and enabling integration with existing enterprise governance frameworks.

\textbf{Unified Treatment of Heterogeneous Participants}: ODP-EL communities naturally accommodate coordination among LLM Agents (task-specific components), Agentic AI (autonomous reasoning systems), and human actors (domain experts, approvers, supervisors) within a single formal framework. This unified treatment proves essential for enterprise agentic AI where human oversight and AI autonomy must coexist within governed structures.

Our design pattern catalogue (Section \ref{sec:catalogue}) instantiates proven solutions as ODP-EL community templates. Individual patterns specify reusable roles, policies, and contracts; pattern compositions define complete community architectures. This approach combines the practical value of design patterns with the formal rigor of ODP-EL, enabling architects to build systems that are both implementable and formally verifiable.

The remainder of this section establishes the formal foundations enabling rigorous agentic AI architecture. Section \ref{subsec:formal-foundations} introduces ODP-EL community concepts and their adaptation to agentic AI systems. Section \ref{subsec:adapting-patterns} is about adapting patterns to Agentic AI. Section \ref{subsec:three-layer-architecture} transforms the pattern-based 
architecture from Section \ref{sec:demonstration} into complete ODP-EL 
community specifications with explicit roles, deontic constraints, contracts, 
and verifiable governance properties.

\subsection{Formal Foundations}
\label{subsec:formal-foundations}

In developing this design pattern framework we have found that many interactions involving Agentic AI need to be positioned as part of an enterprise environment in which they are to be deployed, either on their own or in collaboration with other Agentic AI\footnote{Recall that LLM Agents can be regarded as simplified form of Agentic AI} or human actors. In order to represent key properties of an enterprise environment, we leverage the ODP EL formalism \cite{IS15414}, which provides a precise way of expressing the enterprise constraints that need to be respected by all actors, Agentic AI and humans, as well as the expression of design variability of the environment\footnote{These variability points are captured through the specification of a policy in ODP-EL, but this description is beyond the scope of this paper; for more details see \cite{linington2025dsl}}, so it can support system changes.

The key concept in ODP-EL is that of \textit{community} \cite{linington2011odp}, 
which is a specification of collaboration rules for multiple roles aimed at 
realizing a shared community objective. A community binds participants 
(computational agents or humans) through their acceptance of defined role 
specifications, including behavioral expectations and normative constraints 
such as obligations, permissions, and prohibitions, also known as deontic constraints. The community also defines interaction protocols in which these roles are involved, as part of the community contract.

When considering Agentic AI architectures, an actor can be an Agentic AI software entity or a human or an organisation. This grounding in the ODP community formalism, which in its own right can be regarded as an organisational pattern, ensures architecture rigor, enables verification, and provides theoretical foundations for reasoning about system properties.

\subsection{Adapting Design Patterns to Agentic AI}
\label{subsec:adapting-patterns}

Agentic AI systems introduce several characteristics that require pattern adaptation beyond traditional software engineering:

\textbf{Unpredictability}: Agentic AI systems exhibit unpredictability from 
two sources. First, the underlying LLM layer introduces stochastic 
non-determinism through probabilistic token sampling, where identical inputs 
may yield different outputs. Second, agentic behavior adds adaptive 
unpredictability through learning from experience, dynamic strategy adjustment, 
and emergent multi-agent interactions. Patterns must account for both sources: 
implementing guardrails for stochastic variation while enabling beneficial 
adaptation and constraining harmful emergent behaviors.\footnote{The ODP-EL 
community has explored modal semantics over possible worlds for reasoning about 
obligations and permissions across non-deterministic system evolutions 
(ISO/IEC 15414 Annex C, informative). While not yet part of the normative 
standard, such formalisms provide promising foundations for verifying governance 
properties despite unpredictable agentic behaviors.}

\textbf{Emergent Behavior}: Multi-agent coordination can produce unexpected collective behaviors not predictable from individual agent specifications. Patterns need mechanisms for monitoring emergence, constraining undesirable behaviors, and leveraging positive emergence for system benefits.

\textbf{Intent and Accountability}: Agentic systems operate based on goals and intentions rather than just procedures. Patterns must support intent modeling (goal + plan + commitment), delegation of obligations with traceability, and accountability tracking---particularly critical in regulated domains where provenance must be maintained.

\textbf{Learning and Adaptation}: Unlike static software, agentic systems learn from experience and adapt behavior over time. Patterns must address memory management across sessions, feedback incorporation mechanisms, and continuous improvement strategies while preventing catastrophic forgetting or bias amplification.

\textbf{Human-AI Coordination}: Enterprise systems require seamless integration of human expertise, oversight, and decision-making with AI autonomy. Patterns must specify how humans and AI agents coordinate as peers within governed frameworks, when escalation to humans is required, and how to maintain appropriate authority boundaries.

These characteristics necessitate new pattern categories (e.g., Reflexion for self-improvement, Constitutional AI for value alignment, Inter-Agent Communication with deontic tokens for accountability) while adapting classical patterns (e.g., extending Orchestration to handle non-deterministic agents, adapting Error Recovery for probabilistic failures).

Our pattern catalogue remains deliberately model-agnostic. Modern large language models provide the cognitive foundation that, when combined with architectural patterns from our catalogue, enable Agentic AI implementations with genuine agency. Specific models evolve rapidly; architectural patterns endure. The patterns we present apply equally whether implemented using GPT-family models, Claude, Gemini, or future LLM architectures, ensuring the catalogue's continued relevance as foundation models advance.

\subsubsection{Agentic Community as ODP-EL Community}

An Agentic Community can be architected using the ODP-EL community concept\cite{linington2011odp}. This model provides both formal semantics and practical implementation guidance. An ODP-EL community comprises four essential elements:

\textbf{Roles}: Specifications of behavior expressing capability of three role types: (1) \textit{LLM Agent roles}---task-specific entities (extractors, validators, transformers); (2) \textit{Agentic AI roles}---autonomous reasoning agents (planners, learners, reasoners); and (3) \textit{Human roles}---domain experts, approvers, supervisors, decision-makers. 

Roles are abstract specifications; multiple agents can fill the same role at different points in time, or multiple instances of the same role type simultaneously if their cardinality is greater than one; roles can be dynamically assigned based on workload or expertise, as long as role filling conditions are satisfied (e.g. RBAC access policy constraint).

\textbf{Normative constraints}: Rules governing role behavior and the entities fulfilling those roles, expressed using deontic constraints---obligations, permissions, prohibitions and authorizations---encapsulated as deontic tokens. ODP-EL's operational semantics for expressing accountability concepts is centered on such tokens: \textit{Burden} tokens represent obligations that participants must fulfill (e.g., \texttt{burden(verify\_consent, ConsentMgr)}); \textit{Permit} tokens represent permission for specific actions (e.g., \texttt{permit(read\_data, MatchingAgent)}); and \textit{Embargo} tokens represent prohibitions on certain behaviors (e.g., \texttt{embargo(auto\_enroll, ALL\_AI)}). These tokens, through their creation, passing to other entities and deletion, create traceable accountability chains---each action is permitted by a permit token, obligations are tracked through burden tokens, and prohibited actions are enforced through embargo tokens.

\textbf{Community Contract}: Normative relationships between roles specifying mutual obligations, permissions, and prohibitions. Contracts bind roles together and define their coordination protocols. In fact, each community specification is defined by a community contract. Example: A \texttt{PhysicianReviewContract} might specify that an LLM Agent has the burden to provide recommendations but only the physician has the permit to approve enrollment.

The practical implementation of ODP-EL deontic tokens in industrial systems has been explored in collaboration with XMPRO Technologies \cite{ResponsibleDTs}. XMPRO's Multi-Agent Generative Systems (MAGS) adopt these concepts through what they term "Rules of Engagement"---governance frameworks implemented using deontic tokens based on ODP-EL semantics~\cite{milosevic_vanschalkwyk2024}. This early production effort validates the practical viability of formal deontic governance in industrial multi-agent systems.

\subsubsection{Towards Intent Modeling}

We formally characterize intent as comprising three interrelated elements: 
goal (desired state), plan (strategy to achieve the goal), and commitment 
readiness (willingness to pursue the plan) \cite{cohen1990intention}. 
This BDI-inspired characterization can be related to ODP-EL's notion of 
commitment---defined as an action resulting in an obligation 
by one or more participants to comply with a rule or perform a contract 
\cite{IS15414}. The key distinction: while intent involves commitment readiness 
(an internal mental state), ODP-EL commitment manifests as an externally 
observable obligation. This establishes a formal link between internal intent 
and external normative constraints---an agent's internal commitment readiness 
may lead to external commitments that create obligations, but the internal 
state itself remains distinct from and non-transferable to the external 
obligation. This distinction is particularly relevant in multi-agent 
negotiation, where an actor's observable actions may reveal their underlying 
intentions, influencing how other agents respond and coordinate.\footnote{This 
parallels the intention-behavior gap and behavior-perception gap identified 
in human collaboration research \cite{minson2025disagree}, where individuals can only 
infer intentions from observed behaviors rather than accessing internal mental 
states directly---a fundamental constraint that applies equally to human-AI and 
AI-AI coordination.}

Intent represents the internal cognitive state of an actor---whether 
Agentic AI or human. It resides within the individual agent as an 
autonomous mental attitude that drives action selection and strategy 
formulation. Note that intent cannot be transferred or delegated from one 
agent to another. When responsibility is delegated, the receiving agent may 
form their own intent, which may in turn lead to a commitment to fulfill the 
delegated obligation; they cannot simply "inherit" the delegating agent's 
goals, plans, and commitment. This principle preserves the autonomy essential 
to genuine agency: each agent pursues objectives through their own 
reasoning processes.

Obligations, in contrast, can be formally delegated between agents. 
In ODP-EL terminology, obligations are expressed as 
burden tokens that can be transferred between community participants while 
maintaining complete traceability of the delegation chain \cite{IS15414}. 
The action of transferring an obligation is referred to as a speech act 
\cite{linington2025dsl}. The deontic token flows from one actor (the principal) 
to another (the agent), creating an explicit record of responsibility 
transfer.\footnote{This principal-agent terminology follows legal and economic 
agency theory.} Speech acts---special actions that create, transfer, or discharge 
obligations, permissions, or prohibitions---formalize accountability-aware 
communication between actors in ODP-EL. These obligations apply uniformly to 
both Agentic AI actors and human participants, with the same formal semantics 
governing delegation. However, legal responsibility ultimately rests with 
legal entities (termed parties in ODP-EL) who are accountable for actions of 
agents operating on their behalf.

Formally, we can express this as:
\begin{itemize}[leftmargin=*]
\item $\text{Intent}_{\text{agent}}(g, p, c)$ where $g = \text{goal}$, $p = \text{plan}$, $c = \text{commitment\_readiness}$
\item $\text{Commitment}_{\text{ODP-EL}}(\text{agent}, \text{obligation})$ where $\text{obligation} \in \text{Burden\_tokens}$
\item $\forall \text{agent}_i, \text{agent}_j: \text{Intent}_{\text{agent}_i} \not\equiv \text{Intent}_{\text{agent}_j}$ (non-transferability)
\item $\forall \text{obligation}: \text{Can\_delegate}(\text{obligation}, \text{agent}_i, \text{agent}_j)$ (delegability)
\end{itemize}

\textbf{Example}: Consider clinical trial matching. A matching agent has the intent to "find best trials for this patient"---comprising its goal (identifying optimal matches), plan (search and ranking strategy), and commitment (persistent pursuit of this objective). Independently, a physician has the intent to "ensure safe, appropriate enrollment"---their goal (patient welfare), plan (clinical evaluation process), and commitment (excercising professional judgment). The system obligation \texttt{burden(enrollment\_decision, physician)} can be delegated to different physicians as staffing changes, but each physician's intent to ensure patient safety remains their own autonomous cognitive state that cannot be transferred to an AI Agent entity or another physician. The AI provides recommendations following its matching intent, while the 
physician exercises their clinical intent in making final enrollment decisions. 
Neither can assume the other's intent, though both can fulfill delegated  obligations.

This formalization enables rigorous verification of accountability properties in Agentic Communities: we can formally prove that critical decisions have appropriate authority through burden token assignments, trace complete responsibility chains through delegation records showing obligation transfers, and verify that no agent assumes obligations beyond its authorized scope through embargo token constraints.

\subsection{Formal Architecture Specifications Using ODP-EL Communities}
\label{subsec:three-layer-architecture}

Having identified a three-layer solution architecture through pattern 
composition in Section~\ref{subsec:step2-composition}, we now demonstrate 
how such architectures can be formalized as ODP-EL communities. This 
formalization transforms conceptual pattern-based designs into rigorous 
specifications: architectural layers become ODP-EL communities with defined 
roles (fillable by LLM Agents, Agentic AI, or humans), normative constraints 
expressed as deontic tokens (burden, permit, embargo), contracts binding 
roles together, and shared enterprise objects. Using the clinical trial architecture as our example, we illustrate how pattern compositions map to formal specifications with 
operational and verifiable governance properties.

\subsubsection{Layer 1: FHIR Foundation (Data Standards Community)}

This foundation layer provides standardized data access through established healthcare interoperability standards, formalized as the \texttt{DataAccessCommunity}.

\textbf{Patterns Composition}: Structured Extraction (LLM Agent), Access Control (Agentic Community via SMART on FHIR OAuth2), Compliance/Governance (Agentic Community via Consent Resource management).

\textbf{ODP-EL Community Specification}:

\textit{Roles}: FHIRDataProvider, DataExtractionAgent (LLM Agent), ConsentManager (LLM Agent), Patient (human), DataGovernanceOfficer (human)

\textit{Normative Constraints (Policies)}:
\begin{itemize}[leftmargin=*]
\item \texttt{permit(read\_demographics, DataExtractionAgent)} $\wedge$\\
\texttt{burden(verify\_consent, ConsentManager)}---consent required
\item \texttt{embargo(access\_without\_consent, ALL)}---absolute prohibition on unauthorized access
\end{itemize}

\textit{Enterprise Objects}: Shared resources owned by the community and accessible to all role-fillers: ConsentRegistry (tracks patient consent status), AuditLog (immutable access records), PatientDataCache (temporary storage respecting consent boundaries)

Layer 1 interfaces with multiple data sources: patient EHRs via SMART on FHIR, ClinicalTrials.gov API, and FEvIR Platform for FHIR-encoded trial eligibility criteria. This standardization enables interoperability while maintaining security boundaries.

\subsubsection{Layer 2: LLM Agent Workflow (Matching Community)}

Layer 2 implements autonomous reasoning through specialized agents that communicate via structured protocols, formalized as the \texttt{MatchingWorkflowCommunity}.

\textbf{Agents Implementation}:
\begin{itemize}[leftmargin=*]
\item Condition Extraction Agent: Structured Extraction + Semantic Bridge (Agentic AI)
\item Patient Embedding Agent: Memory-Augmented (Agentic AI) + Filtering/Triage
\item Eligibility Structuring Agent: Hierarchical Planning (Agentic AI) + Data Transformation
\item Criteria Matching Agent: ReAct (Agentic AI) + Validation
\item Physician: Human role for clinical decisions
\end{itemize}

\textbf{ODP-EL Community Specification}:

\textit{Roles}: ConditionExtractor (Agentic AI), PatientEmbedder (Agentic AI), EligibilityStructurer (Agentic AI), CriteriaMatcheR (Agentic AI), Physician (human), WorkflowOrchestrator (Agentic AI)

\textit{Normative Constraints (Policies)}:
\begin{itemize}[leftmargin=*]
\item \texttt{permit(evaluate\_eligibility, MatchingAgent)}---\\eligibility assessment permitted

\item \texttt{embargo(final\_decision, ALL\_AI\_AGENTS)}---\\
AI enrollment decisions prohibited

\item \texttt{burden(make\_enrollment\_decision, Physician)}---\\
physician decision obligation

\item \texttt{burden(provide\_explanation, MatchingAgent)}---\\
explanation required
\end{itemize}

\textit{Contracts}: 
\begin{itemize}[leftmargin=*]
\item \texttt{MatchingWorkflowContract}: Coordinates agents through structured message passing with Inter-Agent Communication pattern
\item \texttt{PhysicianReviewContract}: Specifies that agents provide recommendations but only physicians decide (Human-in-the-Loop pattern)
\end{itemize}

\textit{Enterprise Objects}: TrialCandidateSet (shared state of candidate trials), PatientProfile (processed patient information accessible to all matching agents), WorkflowState (coordination state for Orchestration pattern)

This layer demonstrates vertical composition: simple extraction (LLM Agent) enables planning and reasoning (Agentic AI), coordinated through Agentic Community contracts with explicit human authority boundaries.

\subsubsection{Layer 3: Negotiation Community}

Layer 3 handles dynamic coordination with external systems and human stakeholders through language-based negotiation, formalized as the \texttt{NegotiationCommunity}.

\textbf{Negotiation Participants}:
\begin{itemize}[leftmargin=*]
\item AI: Negotiation Coordinator, Capability Discovery, Semantic Bridge, Conflict Resolution, Compliance Validation agents
\item Humans: Trial site coordinators, data governance officers, clinical staff
\item External Systems: Trial sites, EHR systems, laboratories
\end{itemize}

\textbf{ODP-EL Community Specification}:

\textit{Roles}: NegotiationCoordinator (Agentic AI), CapabilityDiscoverer (Agentic AI), SemanticBridge (Agentic AI), ConflictResolver (Agentic AI), ComplianceValidator (Agentic AI), TrialSiteCoordinator (human), DataGovernanceOfficer (human), ExternalSystem (computational)

\textit{Normative Constraints (Policies)}:
\begin{itemize}[leftmargin=*]
\item \texttt{burden(validate\_compliance, ComplianceAgent)} \\before external communications---regulatory compliance required

\item \texttt{burden(approve\_novel\_request, DataOfficer)} for new data types---\\
human approval for exceptional requests

\item \texttt{permit(negotiate\_protocol, NegotiationCoordinator)}---\\
protocol negotiation authorized

\item \texttt{embargo(share\_PHI\_externally, ALL)} unless\\
\texttt{permit(share\_specific\_data, DataOfficer)} granted---PHI protection
\end{itemize}

\textit{Contracts}:
\begin{itemize}[leftmargin=*]
\item \texttt{NegotiationProtocol}: Implements A2A-style agent-to-agent negotiation using speech acts (propose, accept, reject, counter-propose)
\item \texttt{ExternalSystemNegotiation}: Extends NegotiationProtocol to include human approvers as coordinated participants
\item \texttt{EscalationContract}: Defines when agents must escalate to humans based on confidence thresholds or policy violations
\end{itemize}

\textit{Enterprise Objects}: NegotiationHistory (tracks all negotiation interactions for Audit Trail pattern), CapabilityRegistry (discovered capabilities of external systems), SemanticMappings (terminology translations maintained by Semantic Bridge pattern)

Layer 3 exemplifies Agentic Community patterns at scale: Multi-Agent System, Orchestration, Negotiation, Semantic Bridge, and Compliance/Governance all operate in concert with both AI and human participants. The Audit Trail pattern captures all interactions for accountability, while deontic tokens ensure actions respect authorization boundaries.

The intelligence exhibited by Agentic Communities emerges primarily from 
coordination mechanisms including structured protocols, conversational 
negotiation, and formal governance frameworks. While LLMs provide the cognitive 
foundation for Agentic AI entities, system-level intelligence derives from 
orchestration protocols, shared enterprise objects, normative constraints 
governing interactions, and contract-based coordination---the architectural 
elements that enable multiple participants to achieve outcomes beyond individual 
agent capabilities. This coordination-centric view explains why formal governance 
frameworks prove essential: they specify the coordination mechanisms that 
generate emergent intelligence.

\subsection{Formal Properties and Runtime Verification}
\label{subsec:formal-properties}

The ODP-EL community model enables formal verification of key properties through deontic token analysis:

\textbf{Safety Property}: ``No patient data accessed without consent''
$$\forall a: \texttt{permit(access\_data, a, p)} \rightarrow \exists c: \texttt{burden(consent, p)} \text{ DISCHARGED}$$

This property states that any permit token authorizing data access must have a corresponding discharged burden token proving consent was obtained. Runtime verification monitors permit token issuance and checks consent discharge status.

\textbf{Authority Property}: ``Physicians make enrollment decisions''
$$\forall \text{enrollment}: \texttt{burden(make\_decision, Physician)} \text{ REQUIRED}$$

This property ensures enrollment decisions always require discharging a burden token held by a physician role. The embargo token \texttt{embargo(final\_decision, ALL\_AI\_AGENTS)} provides complementary prohibition-based enforcement.

\textbf{Prohibition Property}: ``AI cannot auto-enroll patients''
$$\forall \text{ai} \in \text{AI\_AGENTS}: \texttt{embargo(final\_decision, ai)} \text{ HOLDS}$$

This property verifies that embargo tokens preventing AI enrollment remain active. Any attempt by AI agents to discharge enrollment burdens triggers violation detection.

\textbf{Accountability Property}: ``All actions traceable to responsible parties''
$$\forall \text{action}: \exists \text{chain}: \texttt{permit\_chain(action)} \wedge \texttt{principal(chain)} \text{ IDENTIFIED}$$

This property ensures every action traces through permit token delegation chains to an identifiable principal (human or organizational entity) ultimately responsible. The Audit Trail pattern maintains these chains for compliance reporting.

These properties are provable from the community specifications and verifiable at runtime through token monitoring

\textbf{Formal Verification Benefits.}
The availability of formally verifiable aspects of solutions are important for 
production-ready Agentic Communities, addressing enterprise requirements.

\paragraph{Practical Business Value.} The ODP-EL formal foundation delivers 
tangible enterprise benefits in regulated industries where autonomous AI failures 
create legal liability, financial risk, or safety hazards: 
(1) Risk Mitigation---provable safety properties reduce legal liability for 
autonomous systems making consequential decisions, establishing defensible 
positions in regulatory audits; 
(2) Continuous Automated Governance---deontic tokens enable governance to execute 
at machine speed, transforming compliance from quarterly reviews to real-time 
continuous control systems; 
((3) Regulatory Approval---formal proofs expedite regulatory review in domains like 
healthcare (HIPAA~\cite{hipaa1996}) and finance (SEC/FINRA~\cite{sec_finra}) 
where autonomous AI requires approval before deployment; and
(4) Operational Confidence---runtime verification enables early detection of 
policy violations before they create compliance incidents.

\paragraph{Verification as Engineering Tool.} Enterprise architects maintain 
diverse verification approaches suited to different assurance levels. For agentic 
AI systems, this engineering palette expands by pattern classification: LLM Agent 
utility patterns require only traditional testing; Agentic AI reasoning patterns 
benefit from behavioral testing and evaluation benchmarks; while Agentic Community 
governance patterns coordinating multiple AI agents and human participants with 
regulatory requirements demand formal verification to prove accountability 
properties that testing cannot adequately validate. This selective formalization---
applying rigorous semantics to governance patterns while maintaining narrative 
documentation for utility patterns---enables both practical implementation and 
verifiable compliance.
\subsection{Industrial Validation: Trust Through Transparency}
\label{subsec:trust-transparency}

Industrial operations face a critical challenge absent in enterprise software: the inability to iterate quickly when autonomous decisions have physical consequences. Equipment failures, safety incidents, and regulatory violations cannot be "rolled back" through software updates. This fundamentally changes the deployment model from assuming autonomy and iterating to fix problems, toward earning trust through transparency before granting autonomy, as discussed comprehensively in \cite{vanschalkwyk2025}

The formal foundations presented in this section directly address these trust 
requirements. ODP-EL's deontic tokens provide machine-verifiable accountability 
chains essential when autonomous decisions carry physical consequences. The 
community specification framework enables explicit representation of authority 
boundaries, ensuring AI agents and human participants coordinate within 
governed structures where responsibility remains traceable. Formal verification 
of safety and authority properties---demonstrated in our clinical trial matching 
case study---becomes not merely theoretical rigor but practical necessity for 
regulatory acceptance and operational confidence in safety-critical domains.

This trust requirement shapes architectural decisions for Agentic Communities 
coordinating AI agents and human participants. First, 
systems must provide dual-purpose infrastructure serving both agent intelligence 
(precedent retrieval, autonomous decisions) and human oversight (complete 
reasoning chains, pattern validation) simultaneously. Second, autonomy should 
expand progressively through deployment modes---advisory (agent recommends, 
human approves), supervised (agent acts within bounds, human monitors), and 
autonomous (agent acts, human reviews)---where trust accumulates through 
validated outcomes rather than being assumed upfront. Third, separation of 
control ensures agents observe, plan, and decide, while a separate control 
layer enforces constraints and executes approved actions, blocking unauthorized 
or prohibited actions regardless of agent recommendations \cite{vanschalkwyk2025}. These principles 
apply equally whether coordinating multiple autonomous agents or integrating 
AI capabilities with human expertise within governed frameworks.

Some early production deployment efforts support these principles. For example, XMPRO's DecisionGraph 
combines Memory-Augmented (\#3), Audit Trail (\#20), Reflexion (\#6), and 
Human-in-the-Loop (\#40) patterns to implement dual-purpose architecture 
where agents retrieve precedents autonomously while humans query complete 
reasoning chains. The Inter-Agent Communication (\#12) and Orchestration 
(\#14) patterns naturally support separation of control through role-based 
specifications, maintaining complete audit trails showing both agent reasoning 
and applied constraints. This architectural approach enables trust to expand 
incrementally as decision traces demonstrate validated reasoning across 
safety-critical operations in asset-intensive industries.

%
%
\section{Related Work}
\label{sec:related-work}

\textbf{Design Patterns and Verification Engineering}: Software engineering progressively adopts rigorous verification as system criticality increases. Aviation software employs DO-178C formal methods \cite{rtca2011}; medical devices follow IEC 62304 verification requirements \cite{iec62304}; automotive systems adopt ISO 26262 safety analysis \cite{iso26262}. These safety-critical domains learned that architectural guidance alone---while valuable for system design---cannot satisfy verification demands when failures create liability. The seminal design patterns work by Gamma et al. \cite{gamma1995design} established narrative documentation of recurring solutions, subsequently extended to distributed systems \cite{hohpe2003enterprise} and enterprise integration. This descriptive methodology proved adequate for traditional software systems amenable to conventional testing.

Agentic AI systems enter this same safety-critical territory. When AI agents make consequential decisions affecting patient safety, financial transactions, or legal compliance, engineering discipline demands verification rigor matching traditional critical systems. Our work extends classical design patterns with verification capabilities drawn from ODP-EL \cite{linington2011odp}, following the precedent that high-stakes software requires both architectural guidance AND verifiable properties. Importantly, we apply selective formalization: LLM Agent utility patterns maintain traditional narrative documentation adequate for conventional testing, while Agentic Community governance patterns employ formal semantics enabling runtime verification, provable compliance, and traceable accountability essential for regulated deployment. 
Our prior work on domain-specific languages for ODP-EL \cite{linington2025dsl} enables domain experts to express governance policies in their terminology while maintaining formal machine-checkable specifications, bridging the gap between regulatory experts and formal verification systems.

\textbf{Agent-Based Systems}: Classical multi-agent systems research \cite{wooldridge2009introduction} established foundations for agent autonomy and coordination. However, these frameworks predate LLMs. Recent work on LLM-based agents \cite{wang2024survey,xi2023rise,weng2023llm} 
explores specific capabilities including memory augmentation~\cite{packer2023memgpt}, 
tool use~\cite{schick2023toolformer}, multi-agent orchestration~\cite{wu2023autogen}, 
embodied reasoning~\cite{ahn2022can}, and value alignment~\cite{bai2022constitutional}, 
but lacks systematic formal foundations for enterprise deployment.

\textbf{Formal Methods for Agents}: Our work builds on BDI models \cite{rao1995bdi}, multi-agent systems foundations~\cite{stone2000multiagent}, agent communication languages \cite{fipa2002acl,colombetti2000commitments}, and deontic logic \cite{hansen2008normative}. The ODP-EL framework \cite{linington2011odp} provides operational semantics we adopt for governance. We extend these with formal intent modeling~\cite{cohen1990intention}, metacognitive reasoning~\cite{cox2011metareasoning}, and human-AI coordination.

\textbf{Enterprise AI}: Recent work on enterprise AI patterns \cite{washizaki2022systematic} focuses primarily on ML operations. Our catalogue extends this to agentic communities requiring autonomy, multi-participant coordination, and formal governance. The ODP-EL grounding distinguishes our work through rigorous formal semantics and verifiable properties.

%
%
%
%

\section{Conclusions and Future Directions}
\label{sec:conclusion}

\subsection{Conclusions}
This paper presents a comprehensive design pattern catalogue for architecting 
agentic AI applications, organized within a three-tier classification framework 
distinguishing LLM Agents, Agentic AI, and Agentic Communities. To enable 
enterprise deployment in regulated industries, we ground Agentic Community 
patterns in ODP Enterprise Language (ODP-EL) formalism, providing verifiable 
governance and accountability through explicit specification of roles fillable 
by AI agents, Agentic AI entities, and human participants; policies expressed 
through deontic logic (obligations, permissions, prohibitions); and contracts 
defining normative relationships. This formal grounding enables implementation 
and verification of applicable policies essential for enterprise deployment 
while supporting the dynamic evolution characteristic of AI systems.

Our classification reveals that nearly half of the patterns represent genuine 
agentic AI with autonomous reasoning capabilities, while over a quarter address 
community-level coordination---reflecting the reality that production systems 
require sophisticated multi-agent orchestration beyond simple task automation. 
The clinical trial matching case study validates our approach by demonstrating 
pattern composition across all three tiers with formal community specifications. 
Token-based accountability through burden, permit, and embargo constructs 
enables runtime verification of compliance properties, while the formal intent 
relationship with externally visible behaviour and related obligation rules 
clarifies accountability chains when AI agents and humans collaborate under 
governed frameworks. This combination of formal rigor and practical patterns 
bridges the gap between autonomous AI capabilities and enterprise deployment 
requirements.

As agentic AI systems transition from research prototypes to production 
environments, the need for systematic architectural guidance grounded in formal 
methods becomes critical. Our framework provides this through a unique 
integration of enterprise architecture standards, design pattern methodology, 
and explicit human-AI coordination models, ensuring systems can meet stringent 
enterprise requirements for governance, accountability, formal verification, 
and regulatory compliance while remaining flexible enough to accommodate the 
dynamic, non-deterministic nature of LLM-based agents and autonomy properties 
of Agentic AI entities.

We emphasize that this catalogue is deliberately model-agnostic. Modern large 
language models provide the cognitive foundation that, when combined with 
architectural patterns from our catalogue, enable Agentic AI implementations 
with genuine agency. Specific models evolve rapidly; architectural patterns 
endure. The patterns presented here apply across LLM implementations, ensuring 
the catalogue's continued relevance as foundation models advance while 
organizational architectures and governance requirements remain stable.

A distinguishing contribution of this work lies in providing formal semantics 
for Agentic Community patterns, addressing a critical gap in design pattern 
methodology. While classical design patterns \cite{gamma1995design,
hohpe2003enterprise} provide invaluable architectural guidance through 
narrative descriptions, they cannot satisfy the verification requirements of 
regulated industries deploying Agentic AI entities. Our selective formalization---
applying rigorous semantics specifically to Agentic Community governance 
patterns while maintaining traditional narrative documentation for utility 
patterns---enables the prototype-to-production transition by providing 
provable safety properties, verifiable compliance, and traceable 
accountability chains that auditors, regulators, and insurance underwriters 
demand. This formal foundation proves essential not as academic exercise but 
as practical necessity for enterprise deployment in domains where autonomous 
AI failures create legal liability, financial risk, or safety hazards.

\subsection{Future Directions}
\label{subsec:future-directions}

We view this work as establishing foundational principles for agentic AI 
architecture while providing practical implementation guidance through our 
pattern catalogue. The catalogue is designed to evolve with the field, 
accommodating new patterns and coordination mechanisms as agentic AI 
technologies mature. The formal foundation provides stable semantics while 
the pattern collection grows. By establishing clear formal foundations and 
classification criteria, we create a sustainable framework for capturing 
and disseminating architectural knowledge. Several promising directions emerge 
from this foundation:

\textbf{Usage Patterns and Operational Practices.} Beyond architectural design 
patterns, capturing recurring operational practices for deploying, monitoring, 
and evolving agentic systems in production would provide valuable guidance for 
practitioners. Usage patterns would address progressive autonomy introduction 
(gradually transitioning from human-supervised to autonomous operation), 
human-in-the-loop escalation protocols (defining when agents should defer to 
human judgment), and agent team composition strategies (determining optimal 
team size and specialization). Such patterns complement our design catalogue 
by addressing the deployment and operations of enterprise agentic AI systems, 
particularly relevant as organizations transition from pilot implementations 
to production-scale deployments.

\textbf{Formal Verification and Tooling.} The ODP-EL formal foundation enables 
rigorous verification of pattern implementations through model checking for 
Compliance/Governance patterns using deontic logic, formal verification of 
Access Control policies, automated validation of Audit Trail completeness, 
and proving correctness properties of Negotiation protocols. Complementary 
tooling development approaches \cite{linington2025dsl} can bridge formal specifications 
with domain-expert expressibility through domain-specific languages (DSL), enabling 
practitioners to express governance policies in domain terminology while 
maintaining machine-checkable specifications. Essential tooling includes 
pattern selection decision trees, code generators for community templates 
that instantiate ODP-EL specifications, testing frameworks for multi-participant 
systems, and visualization tools for pattern composition analysis.

\textbf{Domain-Specific Instantiations.} While our patterns provide general 
architectural guidance, domain-specific instantiations would accelerate 
adoption across industries. Specialized pattern collections for healthcare 
(FHIR integration \cite{fhir2023}, clinical decision support), finance (algorithmic trading, 
regulatory compliance), manufacturing (supply chain optimization, quality 
control), and cybersecurity (threat detection, incident response) would 
provide practitioners with tailored starting points aligned with industry-specific 
requirements and standards. These domain templates would preserve the formal 
foundations while addressing sector-specific governance and operational constraints.

\textbf{Community Evolution and Contribution.} The pattern catalogue is 
designed as a living document to accommodate emerging patterns as agentic AI 
capabilities evolve. While the supplementary material accompanying this paper 
presents a reviewed snapshot of 46 validated patterns, we plan to 
maintain the catalogue as a community resource on GitHub, enabling practitioners 
and researchers to contribute new patterns, refinements, and validation cases. 
The paper's supplementary material represents the peer-reviewed baseline; the 
GitHub repository will serve as the canonical reference for ongoing pattern 
development. This includes establishing community contribution processes with 
clear validation criteria, tracking adoption metrics across production 
deployments, and implementing automated consistency checking to ensure new 
patterns align with the established classification framework and formal 
foundations.

These future directions extend the foundational work presented here, supporting 
the transition of agentic AI systems from research innovation to enterprise 
deployment while maintaining the formal rigor essential for regulated industries. 
By combining practical patterns with verifiable formal semantics, we enable 
practitioners to build production-grade agentic AI systems that are both 
powerful and properly governed.

%
%

\subsubsection{\ackname}
This work was supported by UNSW Sydney, under a 
grant obtained by Professor Fethi Rabhi. The first author would like to thank 
Pieter van Schalkwyk (XMPRO Technologies) for many discussions on industrial 
applications and how the value of ODP-EL architecture and governance frameworks 
can support multi-agent environments. We also wish to thank Dr Basem Suleiman and Mr Chinmay Manchanda for their contributions to this project. Systematic documentation and structural 
organization of the pattern catalogue was conducted in collaboration with 
Claude (Anthropic). All conceptual frameworks, formal specifications, and 
domain expertise represent the work of the named authors.

\bibliographystyle{splncs04}
\bibliography{references}

\end{document}